\documentclass[twoside]{article}

\usepackage[accepted]{aistats2020}
\usepackage[round]{natbib}

\usepackage{hyperref}       
\usepackage{url}            
\usepackage{booktabs}       
\usepackage{amsfonts}       
\usepackage{nicefrac}       
\usepackage{microtype}      
\usepackage{multirow}
\usepackage{xcolor,colortbl}
\usepackage{xr}

\usepackage{algorithm}
\usepackage[noend]{algpseudocode}

\usepackage{amsmath, amsfonts, mathtools}

\usepackage{enumitem}
\setitemize{noitemsep,topsep=0pt,parsep=0pt,partopsep=0pt,leftmargin=*}

\def\a{{\mathbf{a}}}
\def\x{{\mathbf{x}}}
\def\z{{\mathbf{z}}}
\def\w{{\mathbf{w}}}

\makeatletter
    \def\blfootnote{\xdef\@thefnmark{}\@footnotetext}
\makeatother

\DeclareMathOperator*{\argmax}{arg\,max}

\makeatletter
\newcommand{\printfnsymbol}[1]{%
  \textsuperscript{\@fnsymbol{#1}}%
}
\makeatother

\let\oldnl\nl
\newcommand{\nonl}{\renewcommand{\nl}{\let\nl\oldnl}}

\begin{document}

%

%

\twocolumn[

\aistatstitle{Robust Variational Autoencoders for Outlier Detection and Repair
of Mixed-Type Data}

\aistatsauthor{ Sim\~ao Eduardo\printfnsymbol{1}$^1$ \And Alfredo Naz\'abal\printfnsymbol{1}$^2$ \And  Christopher K. I. Williams$^{12}$ \And Charles Sutton$^{123}$}

\aistatsaddress{$^1$School of Informatics,
University of Edinburgh, UK \\ $^2$The Alan Turing Institute, UK; \hspace{0.5cm}$^3$Google Research}

]

\blfootnote{\printfnsymbol{1} Joint first authorship.}

\begin{abstract}
We focus on the problem of unsupervised cell outlier detection and repair in
mixed-type tabular data. Traditional methods are concerned only with detecting which rows in the dataset are
outliers. However, identifying which cells are corrupted in a
specific row is an important problem in practice, and the very first step
towards repairing them. We introduce the Robust Variational
Autoencoder (RVAE), a deep generative model that learns the joint
distribution of the clean data while identifying the outlier cells, allowing their imputation (\textit{repair}). 
RVAE explicitly learns the probability of each cell being an outlier, balancing different
likelihood models in the row outlier score, making the method suitable
for outlier detection in mixed-type datasets.
We show experimentally
that not only RVAE performs better than several state-of-the-art methods in
cell outlier detection and repair for tabular data, but also that is robust against the
initial hyper-parameter selection.
\end{abstract}

\section{Introduction}

The existence of outliers in real world data is a problem data scientists face daily, so outlier detection (OD) has been extensively studied in the literature~\citep{chandola2009anomaly, emmott2015meta, hodge2004survey}.
The task is often \emph{unsupervised}, meaning that
we do not have annotations indicating whether individual cells
in the data table are clean or anomalous.
Although supervised OD algorithms
have been proposed~\citep{Lee2018TrainingCC, an2015variational, schlegl2017unsupervised}, annotations of anomalous cells are
often not readily available in practice.
Instead, unsupervised OD attempts to infer
the underlying clean distribution, and explains outliers as instances that deviate from that distribution.
It is important
to focus on the joint distribution over features,
because although some outliers can be easily identified as anomalous
by considering only the marginal distribution of the feature itself, many others are only detectable within the context of the other features~\citep[section 2.2]{chandola2009anomaly}.
Recently deep models have outperformed
traditional ones for tabular data tasks~\citep{Klambauer2017SelfNormalizingNN}, capturing their underlying structure better. They are an attractive choice for OD, since they have the flexibility to model a wide variety of clean distributions.
However, OD work has mostly focused on image datasets, 
repairing dirty pixels instead of cells in tabular data~\citep{wang2017green, zhou2017anomaly, akrami2019robust}.

Outliers present unique challenges to deep generative models.
First, most work focuses on detecting anomalous 
data rows, without detecting which specific cells in a row are problematic~\citep{redyuk2019learning, schelter2018automating}. However, not enough care is given to cell granularity, which means it is often difficult to
properly \textit{repair} the dirty cells, e.g. if there are a large number of columns or 
when the data scientist is not a domain expert.
Work on cell-level detection and repair often focuses on real-valued features, e.g. images~\citep{zhou2017anomaly, wang2017green, schlegl2017unsupervised}, or does not provide a principled way to detect anomalous 
cells \citep{Nguyen2018ScalableAI}.
Second,
tabular data is often \emph{mixed-type}, including both continuous and categorical columns.
Although modelling mixed-type data has been explored before~\citep{nazabal2018handling,vergari2019automatic},
a difficulty arises when handling outliers.
Standard outlier scores are based on the probability that
the model assigns to a cell, but these values are not comparable
between likelihood models, performing poorly for mixed-type data.
Finally,
the effect of outliers in unsupervised learning can be insidious.
Since deep generative models are highly flexible, they are not always robust against outliers~\citep{Hendrycks2019BenchmarkingNN}, overfitting to anomalous cells. When the model overfits, it cannot identify
these cells as outliers, because it has modelled them as part of the
clean distribution, and consequently, most repair proposals are skewed towards the dirty values, and not the underlying clean ones.

Our main contributions are:
(i) \emph{Robust Variational Autoencoder (RVAE)}, a novel fully unsupervised deep generative model for cell-level OD and repair for mixed-type tabular data.
    It uses a two-component mixture model for each feature, with one component for clean data, 
    and the other component that robustifies the model by isolating outliers.
    (ii) \textit{RVAE} models the underlying clean data distribution by
    down-weighting the impact of anomalous cells, providing a competitive 
    outlier score for cells and a superior 
    estimate of cell repairs.
    (iii) A hybrid inference scheme for optimizing the model parameters,
    combining amortized and exact variational updates, which proves superior to standard amortized inference.
    (iv) \textit{RVAE} allows us to present an outlier score that is commensurate across mixed-type data.
    (v) \textit{RVAE} is robust to the selection of its hyper-parameters, 
    while other OD methods suffer from the need to tune their parameters to each specific dataset.

\section{Variational Autoencoders}
\label{sec:VAE}

We consider a tabular dataset $X$ with $n \in \{1,\cdots,N$\} instances and $d \in \{1,\cdots,D\}$ features, where each cell $x_{nd}$ in the dataset can be real (continuous), $x_{nd} \in \mathbb{R}$, or categorical, $x_{nd} \in \{1,..,C_d\}$ with $C_d$ the number of unique categories of feature $d$.

Cells in the dataset are potentially corrupted with an unknown noising process appropriate for the feature type. The objective in this work is not only detecting the anomalous instances in the dataset, termed \emph{row outliers}, but also determining the specific subset of cells that are anomalous, termed \emph{cell outliers}, proposing potential \emph{repair} values for them.

A common approach to unsupervised OD is to build a generative model $p(X)$ 
that models the distribution of clean data. A powerful class of deep generative models
are variational autoencoders (VAEs) \citep{Kingma14}, which model $p(X)$ as
\begin{align}
\label{eq:likelihood_basic_vae}
     p(X) & = \prod_{n=1}^N \int d\z_n \; p(\z_n) p_{\theta}(\x_n|\z_n),
\end{align}
where $p_\theta (\x_{n} | \z_n) = \prod_{d=1}^D p_\theta (x_{nd} | \z_n)$ and $p_\theta (x_{nd} | \z_n)$ is the conditional likelihood of feature $d$,
$\z_n \in \mathbb{R}^{K}$ is the latent representation of instance $\x_n$, 
and $p(\z_n) = \mathcal{N}(\bf 0 , \bf I)$ is an isotropic multivariate Gaussian prior.
To handle mixed-type data, we choose the conditional likelihood  $p_\theta (x_{nd} | \z_n)$ differently for
each feature type.
For real features $p_\theta (x_{nd} | \z_n) = \mathcal{N}(x_{nd} | m_d(\z_n), \sigma_d)$, where each $\sigma_d$ represents the standard deviation of feature $d$ and they are parameters learnt by the model.
For categorical features $p_\theta (x_{nd} | \z_n) = f(\a_d(\z_n))$, where $\a_d(\z_n)$ is an unnormalized vector of probabilities for each category and $f$ is the softmax function. All $m_d(\z_n)$ and $\a_d(\z_n)$ are parametrized by feed-forward networks.

As exact inference for $p_{\theta}(\z_n|\x_n)$ is generally intractable, 
a variational posterior $q_{\phi}(\z_n|\x_n)$ is used; in VAEs this is 
also known as the encoder.
It is modelled by a Gaussian distribution with parameters $\mu(\x_n)$ and $\Sigma(\x_n)$
\begin{equation}
\label{eq:encoder}
    q_{\phi}(\z_n|\x_n) = \mathcal{N}(\z_n|\mu(\x_n),\Sigma(\x_n)),
\end{equation}
where $\phi = \{\mu(\x_n), \Sigma(\x_n)\}$ are feed-forward
neural networks, and $\Sigma(\x_n)$ is a diagonal covariance matrix.
VAEs are trained by maximizing the lower bound on the marginal log-likelihood called the \emph{evidence lower bound (ELBO)},
given by
\begin{align}
    \mathcal{L} = \frac{1}{N}
    \sum_{n=1}^N \sum_{d=1}^D \mathbb{E}_{q_\phi(\z_n | \x_n)} \left[ \log p_\theta (x_{nd} | \z_n) \right] \nonumber\\ -
    D_{KL}(q_\phi(\z_n|\x_n) || p(\z_n)),
\end{align}
where the neural network parameters of the decoder $\theta$ and encoder $\phi$ are learnt with a gradient-based optimizer.
When VAEs are used for OD, typically an instance in a tabular dataset is declared an outlier if the expected likelihood $\mathbb{E}_{q_\phi(\z_n | \x_n)} \left[\log{p_\theta (\x_{n} | \z_n)}\right]$ is small~\citep{an2015variational, wang2017green}.

\section{Robust Variational Autoencoder (RVAE)}
\label{sec:RVAE}

To improve VAEs for OD and repair, we want to make them more robust by automatically identifying potential outliers during training,
so they are down-weighted when training the generative model. 
We also want a cell-level outlier score which is comparable across continuous and categorical attributes.
We can achieve both goals by modifying the generative model.

We define here our {robust variational autoencoder (RVAE)}, a deep generative model based on a two-component mixture model likelihood (decoder) per feature, which isolates the outliers during training. 
RVAE is composed of a clean component $p_\theta (x_{nd} | \z_n)$ for each dimension $d$, explaining the clean
cells, and an outlier component $p_0(x_{nd})$, explaining the outlier cells.
A mixing
variable $w_{nd} \in \{0,1\}$ acts as a gate to determine whether cell $x_{nd}$ should be modelled by the clean component 
$(w_{nd} = 1)$
or the outlier component $(w_{nd} = 0)$.
We define the marginal likelihood of the mixture model model under dataset $X$ as%
\footnote{Mixture models can also 
be written in product form using mixing variables $w_{nd}$~\citep[Section 9, page 431]{bishop2006pattern}, as we adopt here.}
\begin{align}
    p(X) = \prod_{n=1}^N \sum_{\w_n} \int d\z \;  p(\z_n) p(\w_n) p(\x_n|\z_n,\w_n), \\
    p(\x_n|\z_n,\w_n) = \prod_{d=1}^D p_\theta (x_{nd}|\z_n)^{w_{nd}}p_0(x_{nd})^{1-w_{nd}}, \label{eq:log_like_RVAE}
\end{align}
where $\w_n \in \{0,1\}^{D}$ is modelled by a Bernoulli distribution
$p(\w_n) = \prod_{d=1}^D \text{Bernoulli}(w_{nd}|\alpha),$
and $\alpha \in [0, 1]$ is a parameter that reflects our belief about the cleanliness of the data.
To approximate the posterior distribution $p(\z, \w  | \x),$
we introduce the variational distribution
\begin{equation}
    \label{eq:var_dist_rvae}
    q_{\phi,\pi}(\w,\z | \x) = \prod_{n=1}^N q_{\phi}(\z_n| \x_n) \prod_{d=1}^D q_{\pi}(w_{nd}| \x_n),
\end{equation}
with $q_\phi(\z_n| \x_n)$ defined in \eqref{eq:encoder} and $q_{\pi}(w_{nd} | \x_n) = \text{Bernoulli}(w_{nd} | \pi_{nd}(\x_{n}))$. The probability $\pi_{nd}(\x_{n})$ can be interpreted as the predicted 
probability of cell $x_{nd}$ being clean.
This approximation uses the mean-field assumption that $\w$ and $\z$ are conditionally 
independent given $\x$.
Finally, the ELBO for the RVAE model can be written as
\begin{align}
\label{eq:ELBO_RVAE}
    &\mathcal{L} = \frac{1}{N} \sum_{n=1}^N \sum_{d=1}^D \mathbb{E}_{q_{\phi}(\z_n| \x_n)}  \left[
    \pi_{nd}(\x_{n})
     \log p_\theta(x_{nd}|\z_n) \right] \nonumber \\
     &+ \frac{1}{N} \sum_{n=1}^N \sum_{d=1}^D \mathbb{E}_{q_{\phi}(\z_n| \x_n)} \left[(1- \pi_{nd}(\x_{n})) \log p_0(x_{nd}) \right]  \nonumber \\
     &- \frac{1}{N} \sum_{n=1}^N D_{KL}( q_{\phi}(\z_n | \x_n) || p(\z_n)) \nonumber \\
     &- \frac{1}{N} \sum_{n=1}^N D_{KL} (q_{\pi}(\w_n | \x_n) || p(\w_n)).
\end{align}
Examining the gradients of \eqref{eq:ELBO_RVAE} helps 
to understand the robustness property of the RVAE. 
The gradient of $\mathcal{L}$ with respect to the model parameters $\theta$ is given by
\begin{equation}
\label{eq:smoothed_grad_rvae}
    \nabla_\theta \mathcal{L} = \frac{1}{N} \sum_{n=1}^N \sum_{d=1}^D \pi_{nd}(\x_{n}) \mathbb{E}_{q_{\phi}(\z_n| \x_n)} \left[
     \nabla_\theta \log p_\theta(x_{nd}|\z_n)  \right].
\end{equation}
We see that $\pi_{nd}(\x_{n})$ acts as a weight on the gradient.
Cells that are predicted as clean will have higher values of $\pi_{nd}(\x_{n})$, and so their gradients are weighted
more highly, and have more impact on the model parameters.
Conversely, cell outliers with low values of $\pi_{nd}(\x_n)$ will have their gradient contribution down-weighted. 
A similar formulation can be obtained for the encoder parameters $\phi$.

\subsection{Outlier Model}
\label{sec:p0_component}

The purpose of the outlier distribution $p_0(x_{nd})$ is to explain the outlier cells in the dataset, removing their effect in the optimization of the parameters of clean component $p_\theta$.
For categorical features, we propose using the uniform distribution
$p_0(x_{nd}) = {C_d}^{-1}$. Such a  uniform distribution assumption has been used 
in multiple object modelling \citep{williams2003learning} as a way to factor in pixel occlusion. 
In~\citet{Chemudugunta2006ModelingGA} a similar approach for background words is proposed.
For real features, we standardize the features to have mean $0$ and standard deviation $1$.
We use an outlier model based on a broad Gaussian distribution\footnote{This is standard~\citep{Quinn2009FactorialSL, Gales1999TailDM}}
$
    p_0(x_{nd}) = \mathcal{N}(x_{nd}|0,S)
$,
with $S>1$. Anomalous cells modelled by the outlier component will be further apart from $m_d(\z_n)$ relative to clean ones.

Although more complex distributions can be used for $p_0(x_{nd})$, 
we show empirically that these simple distributions are enough to detect outliers from a range of noise levels (Section~\ref{sec:experiments}).
Furthermore, RVAE can easily be extended to handle other types of features~\citep{nazabal2018handling}: for count features we can use a Poisson likelihood, where the outlier component $p_0$ would be a Poisson distribution with a large rate; for ordinal features we could have an ordinal logit likelihood, where $p_0$ can be a uniform categorical distribution.

\subsection{Inference}
\label{sec:inference}

We use a hybrid procedure to train the parameters of RVAE that alternates
amortized variational inference 
using stochastic gradient descent for $\phi$ and $\theta$, and coordinate ascent over $\pi$.
When we do not amortize $\pi,$ but rather treat each
$\pi_{nd}(\x_{n}) \in [0, 1]$ as an independent
parameter of the optimization problem,
then
an exact solution for $\pi_{nd}(\x_{n})$ is possible when $\phi$ and $\theta$ are fixed. Optimizing the ELBO \eqref{eq:ELBO_RVAE} w.r.t. $\pi_{nd}(\x_{n})$, we obtain an exact expression for the optimum\footnote{The derivation of equation~\eqref{eq:exact_pi_fsigma} is provided in the
Supplementary Material (Section 2)}
\begin{align}
    \label{eq:exact_pi_fsigma}
    \hat{\pi}_{nd} (\x_{n})  =  g \left(
    r + \log{\frac{\alpha}{1-\alpha}}\right), \\\nonumber
    r  = \mathbb{E}_{q_\phi(\z_n|\x_n)} \left[ 
    \log{\frac{p_\theta(x_{nd}|\z_n)}{p_0(x_{nd})}} \right],
\end{align}
where $g$ is the sigmoid function.
The first term in \eqref{eq:exact_pi_fsigma} represents the density 
ratio $r$ between the clean component $p_\theta(x_{nd}|\z_n)$ and the outlier component $p_0(x_{nd})$. When  $r>1$ it will bias the decision towards assuming the cell being clean, conversely  $r<1$ it will bias the decision towards the cell being dirty.
The second term in \eqref{eq:exact_pi_fsigma} represents our prior belief about cell cleanliness, defined by $\alpha \in [0,1]$. Higher values of $\alpha$ will skew the decision boundary towards a higher $\hat{\pi}_{nd} (\x_{n})$, and vice-versa.
This coordinate ascent strategy is common
in variational inference for conjugate exponential family distributions (see e.g. \citealt{Jordan1999AnIT}). 
We term this model RVAE-CVI (Coordinate ascent Variational Inference) below.

Alternatively, $\pi_{nd}(\x_{n})$ can be obtained using amortized variational inference.
However, two problems arise in the process. First, an inference gap is introduced by amortization, leading to slower convergence to the optimal solution.
Second, there might not be enough outliers in the data to properly train a neural network to recognize the decision boundary between clean and dirty cells. We term this model RVAE-AVI 
(Amortized Variational Inference). RVAE inference is summarized in Algorithm
\ref{alg:RVAE_inference}, for both
the coordinate ascent version (RVAE-CVI) 
and the amortized version (RVAE-AVI). 
We used Adam~\citep{Kingma2014AdamAM} as the gradient-based optimizer (line 15).
\begin{algorithm}[ht]
\caption{RVAE Inference}
\label{alg:RVAE_inference}
\begin{algorithmic}[1]
\Procedure{RVAE}{$\eta$ learning rate, $M$ batch size, $T$ number epochs, 
$\alpha$ prior value}     
    \If{RVAE-AVI = True}
        \State Define NN parameters: $\Psi = \{\phi, \theta, \tau \}$;
    \ElsIf{RVAE-CVI = True}
        \State Define NN parameters: $\Psi = \{\phi, \theta \}$;
    \EndIf
    \State Initialize $\Psi$;
    \For{$1,...,T$}
        \State Sample mini-batches $\{X_m\}_{m=1}^M \sim p(X)$;
        \State Evaluate $p_\theta(x_{md}|\z_m)$ and $p_0(x_{md})$ $\forall m,d$;
        \If{RVAE-AVI = True}
            \State Evaluate encoder $\pi_\tau(\x_n)$;
        \ElsIf{RVAE-CVI = True}
            \State Infer $\hat{\pi}_{md}, \forall m,d $ using eq. \eqref{eq:exact_pi_fsigma}
        \EndIf
        \State $g_\Psi \xleftarrow{} \nabla_\Psi \mathcal{L}(\Psi, \pi(\x_n), \alpha)$
        using eq. \eqref{eq:ELBO_RVAE};
        \State $\Psi \xleftarrow{} \text{Optimizer} (\Psi, g_\Psi, \eta)$;
    \EndFor
\EndProcedure
\end{algorithmic}
\end{algorithm}

\subsection{Outlier Scores}
\label{sec:outlier_score}

A natural approach to determine which cells are outliers in the data is computing the likelihood of the cells under the trained model. In a VAE, the scores for row and cell outliers would be
\begin{align}
    \text{\bf Cell:} \; & -\mathbb{E}_{q_\phi(\z_n | \x_n)}\left[\log{p_\theta(x_{nd}|\z_n)}\right], \nonumber \\
    \text{\bf Row:} \; & -\sum_{d=1}^D\mathbb{E}_{q_\phi(\z_n | \x_n)}\left[ \log{p_\theta(x_{nd}|\z_n)}\right],\label{eq:vae_outlier_score}
\end{align}
where a higher score means a higher outlier probability. However, likelihood-based outlier scores present several problems, specifically for row scores. In mixed-type datasets categorical features and real features are modelled by probability and density distributions respectively, which have different ranges. Often this leads to continuous features dominating over categorical ones. 
With the RVAE we propose an alternative outlier score based on the mixture probabilities $\hat{\pi}_{nd}(\x_n)$
\begin{equation}
    \text{\bf Cell:} \; -\log \hat{\pi}_{nd}(\x_n), \quad \text{\bf Row:} \; -\sum_{d=1}^D \log \hat{\pi}_{nd}(\x_n),
    \label{eq:pi_outlier_score}
\end{equation}
where again a higher score means a higher outlier probability. Notice that the row score is just the negative log-probability of the row being clean, given by $\hat{\pi}_n = \prod_{d=1}^D \pi_{nd}(\x_n)$. These mixture-based scores are more robust against some features or likelihood models dominating the row outlier score, making them more suitable for mixed-type datasets.

\subsection{Repairing Dirty Cells}

Cell repair is related to missing data imputation. However, this is a much \textbf{harder} task, 
since the positions of anomalous cells are not given, and need 
to be inferred.
After the anomalous cells are identified, a robust generative model allows to impute them
given the dirty row directly.
In general, repair under VAE-like models can be obtained via maximum a posteriori (MAP) inference,
\begin{equation}
\label{eq:AE_repair_method}
     \hat{x}^{i}_{nd} = \argmax_{x_{nd}} p_\theta(x_{nd} | \z_n), \ \ \ \z_n \sim q_\phi(\z_n | \x^o_n),
\end{equation}
where superscript $i$ denotes imputed or clean cells (depending on context), and $o$ corresponds to observed or dirty cells. 
In the case of RVAE, $p_\theta(x_{nd} | \z_n)$ is the clean component 
responsible for modelling the underlying clean data, see~\eqref{eq:log_like_RVAE}.
This reconstruction is akin to robust PCA's clean component.
In practice, for real features $\hat{x}^{i}_{nd}=m_d(\z_n)$, the mean of the Gaussian likelihood, and for categorical features $\hat{x}^{i}_{nd}= \argmax_c f(a_{dc}(\z_n))$, the highest probability category.
Other repair strategies are discussed in the Supplementary Material (Section 10).

\section{Experiments}
\label{sec:experiments}

We showcase the performance of RVAE and baseline methods, 
for both the task of identifying row and cell outliers and repairing the corrupted cells in the data\footnote{\url{https://github.com/sfme/RVAE_MixedTypes/}}.
Four different datasets from the UCI repository \citep{Lichman},
with a mix of real and categorical features, were selected for the evaluation (see
Supplementary Material, Section 1).
We compare RVAE with ABDA~\citep{vergari2019automatic} on a different
OD task in the Supplementary material (Section 9).

\subsection{Corruption Process}
\label{sec:corruption_description}

All datasets were artificially corrupted
in both training and validation sets.
This is a standard practice in OD~\citep{Futami2018VariationalIB,redyuk2019learning,Krishnan2016ActiveCleanID,Natarajan2013LearningWN}, and a necessity in our setting, due to the scarcity of available datasets with labelled cell outliers.
No previous knowledge about corrupted cell position, or dataset corruption proportion is assumed.
For each dataset, a subset of cells are randomly selected for corruption, following a two-step procedure: a) a percentage of rows in the data are selected at random to be corrupted; b) for each of those selected rows, 20\% of features are corrupted at random, with different sets of features being corrupted in each select row. For instance, a 5\%-20\% scenario means that 5\% of the rows in the data are randomly selected to contain outliers, and for each of these rows, 20\% of the features are randomly corrupted, leading to 
1\% of cells corrupted overall in the dataset.
We will consider for the experiments five different levels of row corruption, $\{1\%,5\%,10\%,20\%,50\%\}$, leading to five different levels of cells corrupted across the data, $\{0.2\%,1\%,2\%,4\%,10\%\}$.

\paragraph{Real features:} Additive noise is used as a noising process, with dirty cell values obtained as
$x_{nd}^{o} \sim x_{nd}^i + \zeta$, with $\zeta \sim p_{noise}(\mu, \eta)$. Note that the noising process is performed before standardizing the data.
Four different noise distributions $p_{noise}$ are explored:
\emph{Gaussian noise} ($\mu = 0$, $\eta = 5\hat{\sigma}_d$), with $\hat{\sigma}_d$ the statistical standard deviation of feature~$d$; \emph{Laplace noise} ($\mu=0$, $\eta = \{4\hat{\sigma}_d,8\hat{\sigma}_d\}$); \emph{Log-Normal noise} ($\mu=0$, $\eta = 0.75\hat{\sigma}_d$); and a \emph{Mixture of two Gaussian noise components} ($\mu_1 = -0.5, \eta_1=3\hat{\sigma}_d$, with probability 0.6 and $\mu_2 = 0.5, \eta_2=3\hat{\sigma}_d$ with probability 0.4).

\paragraph{Categorical features:} The noising process is based on the underlying marginal (discrete) distribution. We replace the cell value by a dirty one by sampling from a \emph{tempered categorical distribution}\footnote{Also known as \textit{power heuristic} in importance sampling.} (and excluding the current clean category):
\begin{equation}
\label{eq:cat_noise}
    {x_{ndc}^{o}} \sim \frac{p_c(x_{nd}^i)^{\beta}}{\sum_{c=1}^{C_d}p_c(x_{nd}^i)^{\beta}},
\end{equation}
with the range $\beta = [0,0.5,0.8]$. Notice that, when $\beta=0$, the noise process reduces to the uniform distribution, while when $\beta=1$, the noising process follows the marginal distribution.

\subsection{Evaluation metrics}
\label{sec:eval_metrics}

In the OD experiments, we use Average Precision (AVPR)~\citep{salton1986introduction, Everingham2014ThePV}, computed according to the outlier scores for each method. AVPR is a measure of area under the precision-recall curve, so higher is better. For cell outliers we report the macro average of the AVPR for each feature in the dataset\footnote{The AVPR macro average is defined as the average of the AVPR for all the features in a dataset.}.
In the repair experiments,
different metrics are necessary depending on the feature types.
For real features, we compute the Standardized Mean Square Error (SMSE) between the estimated values $\hat{x}_{nd}^i$ and the original ground truth in the dirty cells $x_{nd}^i$, normalized by the empirical variance of the ground truth values:
$
    SMSE_d = \frac{\sum_{n=1}^{N^d_c} (x_{nd}^i - \hat{x}_{nd}^i)^2}{\sum_{n=1}^{N_c} (x_{nd}^i - \overline{x}_d)^{2}},
$
where $\overline{x}_d$ is the statistical mean of feature $d$ and 
$N^d_c$ is the number of corrupted cells for that feature\footnote{In our experiments $\overline{x}_d = 0$ in practice, since the data has been standardized before using any method}.
For categorical features, we compute the Brier Score between the one-hot representation of the ground truth $x_{nd}^i$ and 
the probability simplex estimated for each category in the feature:
$
    Brier_d = \frac{1}{2N_c}\sum_{n=1}^{N^d_c}\sum_{c=1}^C (x^i_{ndc} - p_c(x^o_{nd}))^2,
$
where $p_c(x_{nd}^o)$ is the probability of category $c$ for feature $d$, $x_{ndc}^i$ the one-hot true value for category $c$, 
and $C$ the number of unique categories in the feature. We used the coefficient $\frac{1}{2}$ in the Brier score to
limit the range to $[0,1]$. We name both metrics as SMSE below for simplicity, 
but the correct metric is always used for each type.

\subsection{Competing Methods}
\label{sec:baselines}

We compare to several standard OD algorithms. Most methods are only concerned about row OD, whilst only a few can be used for cell OD. For more details on parameter selection and network settings for RVAE and competitor methods, see the Supplementary Material (Section 3).
\paragraph{Exclusively row outlier detection.} We consider
\textit{Isolation Forest (IF)} \citep{liu2008isolation}, an OD algorithm based on decision trees,
which performed quite well in the extensive comparison of \citet{emmott2015meta};
and \textit{One Class Support Vector Machines (OC-SVM)} \citep{chen2001one}
using a radial basis function kernel.
\paragraph{Row and cell outlier detection.} We compare to         (\textit{i})
    estimating the
    \textit{Marginal Distribution} for each feature and using the negative log-likelihood as the outlier score. For real features we fit a Gaussian mixture model with the number of components chosen with the Bayesian Information Criterion. The maximum number of components is set at 40. For categorical features, the discrete distribution is given by the normalized category frequency; (\textit{ii}) a combination of \textit{OC-SVM and Marginal Distribution} for each feature. We use Platt scaling to transform the outlier score of OC-SVM for each row (to obtain log-probability), and then combine it with marginal log-likelihood of each feature. This score, a combined log-likelihood, is then used for cell OD;
(\textit{iii}) \textit{VAEs} with $\ell_2$ regularization and outlier scores given by \eqref{eq:vae_outlier_score};
    (\textit{iv}) \textit{DeepRPCA} \citep{zhou2017anomaly}, an unsupervised model inspired by robust PCA. The data $X$ is divided in two parts
    $X = R + S$, where $R$ is a deep autoencoder reconstruction of the clean data, and $S$ is a sparse matrix containing
    the estimated outlier values (see Supplementary Material, Section 3, for further details).
    Outlier scores for rows are given by the Euclidean norm 
    $\sqrt{\sum_{d=1}^D|s_{nd}|^2}$, whilst cell scores are given by $|s_{nd}|^2$, where $s_{nd} \in S$.
(\textit{v}) A set of \textit{Conditional Predictors} (\textit{CondPred}), where a neural network parametrizing $p_\theta(\x_{n})$ is employed for each feature in the data given the rest\footnote{This can be seen as a pseudo-likelihood
model given by $p_\theta(\x_{n}) \approx \prod_d p_\theta(x_{nd} |\x_{n \, \setminus d})$}.
However, $\ell_2$ regularization 
is necessary to prevent overfitting, and the model is overall much slower to train than VAE .

\paragraph{Repair:} We compare to \textit{VAE}, \textit{DeepRPCA}, \textit{Marginal Distribution} method and \textit{Conditional Predictor} (\textit{CondPred}) method for repairing dirty cells 
(same model parameters as in OD). 
We use~\eqref{eq:AE_repair_method} for all VAE-based methods. For DeepRPCA we use $\hat{X}^i = R$. For CondPred the estimate is $\hat{x}^{i}_{nd} = \argmax_{x_{nd}} p_\theta(x_{nd} |\x^o_{n \, \setminus d})$, with $\x^o_{n \, \setminus d}$ meaning all features in $\x^o_n$ except $x^o_{nd}$. The Marginal Distribution method takes
$x^o_{nd}$ and uses as estimate the mean of the closest GMM component in the real line.
For RVAE, results using a different inference strategy (pseudo-Gibbs sampling) are provided in the Supplementary Material~\citep{rezende2014stochastic}.

\subsection{Hyperparameter selection for competing methods}
\label{sec:baselines_parameters}

In order to tune the hyperparameters for the competing methods, we
reserved a validation set with known inlier/outlier labels and
ground truth values. This validation set was {\bf not}
used by the RVAE method. Thus the performance obtained by the
competitor methods is an \emph{optimistic} estimate of their
performance in practice.
Note also that
RVAE-CVI is robust to the selection of its
parameter $\alpha$ in \eqref{eq:exact_pi_fsigma}, as we will show in
Section~\ref{sec:alpha_sweep}.
\begin{figure}
    \centering
    \includegraphics[width=1.0\linewidth]{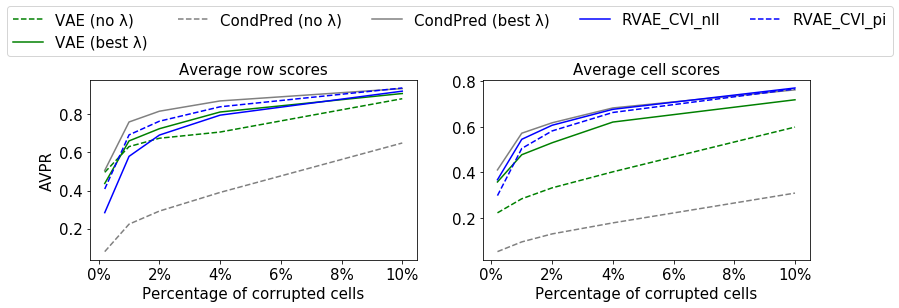}
    \vspace*{-9mm}
    \caption{Row and cell OD performance (higher means better) of VAE and CondPred methods without L2 regularization and best choice of $\lambda$.}
    \label{fig:parameter_selection}
\end{figure}
In Figure~\ref{fig:parameter_selection} we compare the performance of the conditional predictor method and VAE, with respect to RVAE-CVI when $\ell_2$ regularization is not used, and when the best $\ell_2$ regularization value is used for each dataset. We term RVAE-CVI-nll our model with outlier score as defined in~\eqref{eq:vae_outlier_score} and RVAE-CVI-pi our model with outlier score as defined in~\eqref{eq:pi_outlier_score}.
We can observe clearly that a significant gap exists in the performance of these competitor methods when not fine-tuned, making explicit the reliance of these methods on a labelled validation set. In the rest of the experiments we will use the best possible version of each competitor method.

\subsection{Outlier detection}
\label{sec:outlier_detection_simple}

\begin{figure}
    \centering
    \includegraphics[width=1.0\linewidth]{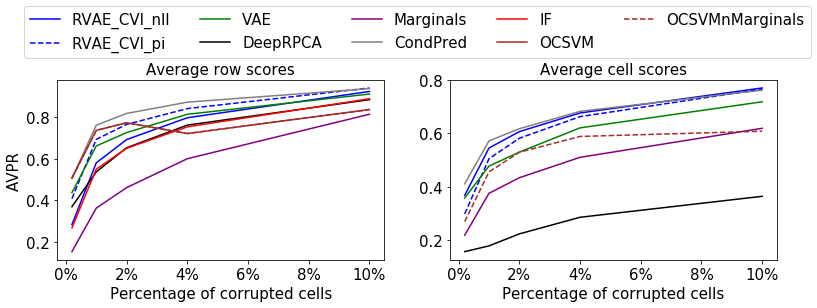}
    \vspace*{-9mm}
    \caption{Row and cell OD scores for the average of the four datasets in 5 different cells corruption levels. Left: AVPR at row level. Right: AVPR at cell level.}
    \label{fig:avpr_average}
\end{figure}

\begin{figure}
    \centering
    \includegraphics[width=1.0\linewidth]{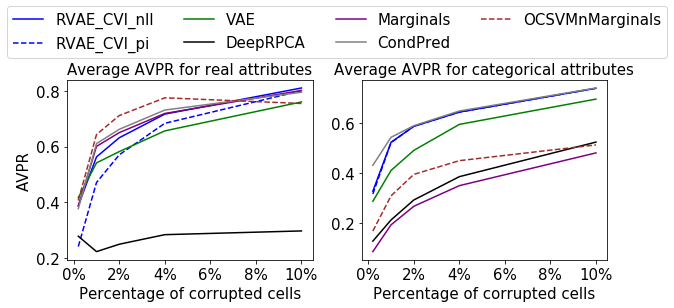}
    \vspace*{-9mm}
    \caption{Average AVPR over all the features in the four datasets partitioned by type. Left: AVPR for real features. Right: AVPR for categorical features}
    \label{fig:avpr_types}
\end{figure}

We compare the performance of the difference methods in OD, both at row and cell levels. We focus on Gaussian noise ($\mu=0, \eta=5\sigma_d$) for real features and uniform categorical noise, i.e.
$\beta=0$ in~\eqref{eq:cat_noise}, relegating results on other noise processes scenarios to Section~\ref{sec:robustness}.
In Figure~\ref{fig:avpr_average} we show the average OD performance across all datasets for all OD models in terms of both row (left figure) and cell OD (right figure).
We relegate RVAE-AVI results to the Supplementary Material (Section 6), since RVAE-AVI is worse than RVAE-CVI in general. Additional results on the OD for each dataset are also available in the Supplementary Material (Sections~4 and 8).
In the right figure, we observe that RVAE-CVI is performing similar to the conditional predictor method on cell OD while being consistently better than the other methods. Additionally, it performs comparatively well in row OD, being similar to the conditional predictor at higher noise levels.
We remind the reader that RVAE-CVI does not need a validation set to select its parameters. This means that RVAE-CVI is directly applicable for datasets where no ground truth is available, providing a comparable performance to other methods where parameter tuning for each dataset is necessary.
Figure~\ref{fig:avpr_average} (left figure) also confirms our hypothesis (Section
\ref{sec:outlier_score}) on the proper score to compute row outliers. We can see in the upper figure that RVAE-CVI using scores based on estimate $\hat{\pi}_{nd}(\x_n)$, as per score~\eqref{eq:pi_outlier_score}, are better for row OD compared to averaging different feature log-likelihoods~\eqref{eq:vae_outlier_score}.
Further analysis of the OD performance of each model for the different
feature types is shown in Figure~\ref{fig:avpr_types}.
While the model based on estimating the marginal distribution works well for real features, it performs poorly on categorical features. Similarly the method combining OCSVM and the marginal estimator detects outliers better than the other methods in real features and low noise levels, but performs poorly for categorical features.
In contrast, RVAE performs comparatively better across different types than the other models, with comparable performance to the conditional predictor.

\subsection{Repair}
\label{sec:repair_experiment}

\begin{figure}
    \centering
    \includegraphics[width=1.0\linewidth]{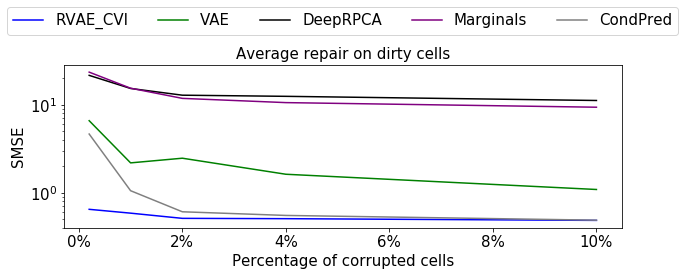}
    \vspace*{-9mm}
    \caption{SMSE computed over the dirty cells in all datasets (lower means better). It shows the average over the four datasets for 5 different noised cells percentages. Y-axis is provided in log-scale.}
    \label{fig:repair_average}
\end{figure}

\begin{figure}
    \centering
    \includegraphics[width=1.0\linewidth]{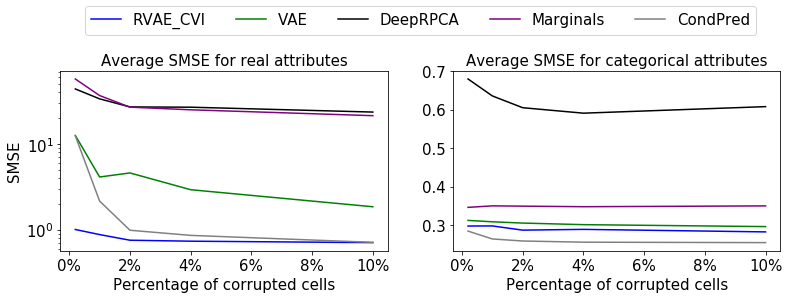}
    \vspace*{-9mm}
    \caption{Average SMSE over all the features in the four datasets according to their type. Left: AVPR for real features. Right: AVPR for categorical features}
    \label{fig:repair_types}
\end{figure}

In this section, we compare the ability of the different models to repair the corrupted 
values in the data. We use the same noise injection process as in Section \ref{sec:outlier_detection_simple}.
Figure~\ref{fig:repair_average} shows the average SMSE repair performance across datasets for all models 
when repairing the dirty cells in the data (more details in the Supplementary Material, Sections 5 and 8).
We can observe that RVAE-CVI outperforms the other models for all the different cell corruption scenarios, being of particularly significance in lower cell corruption regimes. This is significantly important since all the comparator methods required hyperparameter selection and still performed worse than RVAE-CVI.
Also, in Figure~\ref{fig:repair_types} we can see the repair performance of different models according to the types of features in the data.
Notice that RVAE-CVI is consistently better than the other models across real features while being slightly worse on categorical features.

\subsection{Robustness to Noising Processes}
\label{sec:robustness}
\begin{figure}
    \centering
    \includegraphics[width=1.0\linewidth]{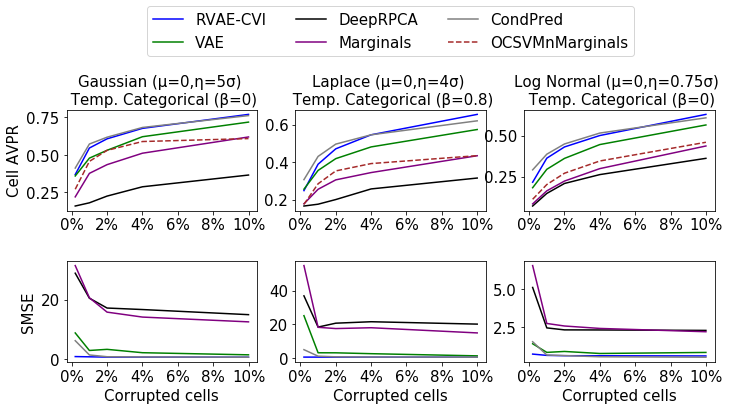}
    \vspace*{-9mm}
    \caption{Effect of three different noising processes.
    Upper figures: average cell OD across datasets. Lower figures: average SMSE on the dirty cells}
    \label{fig:noise_processes}
\end{figure}

Figure~\ref{fig:noise_processes} shows the performance of the different models across different combinations of noise processes for all datasets and noise corruption levels (three other noise processes are covered in the Supplementary Material, Section 7). 
We notice that all the models perform consistently across different types of noise. RVAE-CVI performs better in repair for low-level noise corruption, while providing competitive performance in OD. Also, our choice of outlier models on Section~\ref{sec:p0_component} does not have a negative effect on the ability of RVAE to detect outliers and repair them. Different noise processes define what is feasible to detect and repair.

\subsection{Robustness to hyperparameter values}
\label{sec:alpha_sweep}

\begin{figure}
    \centering
    \includegraphics[width=1.0\linewidth]{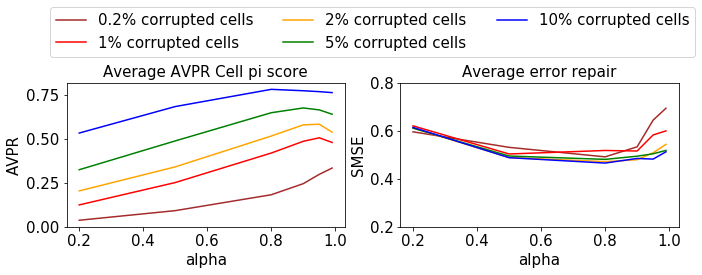}
    \vspace*{-9mm}
    \caption{RVAE-CVI performance with different choices for $\alpha$. Left: average cell AVPR over the datasets. Right: average repair over dirty cells}
    \label{fig:alpha_sweep}
\end{figure}

In this section, we examine the robustness of RVAE inference to the choice $\alpha$, and study its effect in both OD and repair of dirty cells.
We have analyzed values of $\alpha$ in the set $\{0.2,0.5,0.8,0.9,0.99\}$ and evaluated RVAE-CVI in all datasets under all levels of cell corruption and the noising process of Sections~\ref{sec:outlier_detection_simple} and~\ref{sec:repair_experiment}.
Figure~\ref{fig:alpha_sweep} shows the performance of RVAE-CVI in both OD (left figure) and repair (right figure) across different values of $\alpha$.
Larger values of $\alpha$ lead in general to a better OD performance, with a slight degradation when we approach $\alpha=1$. Repair performance is consistent across different choices of $\alpha$, but values closer to 0 or 1 lead to a degradation when repairing dirty cells.

\section{Related Work}

There is relevant prior work 
in the field of OD and robust inference in the presence of outliers,
a good meta-analysis study presented in~\citet{emmott2015meta}.
Different deep models have been applied to this task,
including autoencoders~\citep{Zong2018DeepAG, Nguyen2018ScalableAI, zhou2017anomaly}, VAEs~\citep{an2015variational, wang2017green} and generative adversarial networks~\citep{schlegl2017unsupervised, Lee2018TrainingCC}.
In~\citet{nalisnick2018deep} the authors show that deep models trained on a dataset assign high likelihoods to instances of different datasets, which is problematic in OD.
We identify outliers during training rather than from a fully-trained model, down-weighting their effect on parameter learning. Earlier in training, the model had less chance to overfit, 
so it should be easier to detect outliers.

Most closely related to our model are methods based on
\textbf{robust PCA (RPCA) and autoencoders}.
 They focus on unsupervised learning in the presence of outliers, even though most methods
 need labelled data for hyper-parameter tuning 
\citep{candes2011robust, zhou2017anomaly, Zong2018DeepAG, Nguyen2018ScalableAI, Xu2018UnsupervisedAD, akrami2019robust}. RPCA-based 
alternatives often assume that the features are real-valued, and model the noise
as additive with a Laplacian prior.
A problem in RPCA-type models is that often the hyper-parameter that controls the outlier mechanism is dataset dependent and difficult
to interpret and tune. In \citet{wang2017green}, the authors proposed using a VAE as a recurrent unit, iteratively denoising the images. This iterative approach is reminiscent of the solvers used for RPCA.
However, their work is not easily extended to mixed likelihood models and suffers from the same problems as VAEs when computing row scores (Section~\ref{sec:outlier_score}).

\textbf{Robust Variational Inference.}
Several methods explore robust divergences for variational learning in the presence of 
outliers applied to supervised tasks~\citep{Regli2018AlphaBetaDF, Futami2018VariationalIB}.
These divergences have hyper-parameters which are 
dataset dependent, and can be difficult to tune in unsupervised OD; in contrast, the
$\alpha$ hyperparameter used in RVAE is arguably more interpretable, 
and experimentally robust to misspecification. Recently a VAE model
using one of these divergences in the decoder 
was proposed for down-weighting outliers~\citep{akrami2019robust}. However, in
contrast to our model, they focused on image datasets and are not concerned with cell outliers. The same hyperparameter tuning problem arises,
and it is not clear out to properly extend to categorical features.

\textbf{Bayesian Data Reweighting.} \citet{Wang2017RobustPM} propose an approach
that raises the likelihood of each observation by some weights and then infer both the latent variables and the weights from corrupted data. Unlike RVAE, these weights are only defined for each instance,
so the method
cannot detect cell-level outliers. Also, the parameters of the model are trained via MCMC instead of variational inference,
making them more difficult to apply in deep generative models.

\textbf{Classifier Confidence.} Several methods explore adding regularization to improve neural network classifier robustness to outliers~\citep{Lee2018TrainingCC,Hendrycks2019DeepAD}. 
However, the regularization hyper-parameters are not interpretable and often require a validation dataset to tune them.
Other works like~\citet{Hendrycks2017ABF}, use the confidence of the predicted
distribution as a measure of OD.

\section{Conclusions}

We have presented RVAE, a deep unsupervised model for cell outlier
detection and repair in mixed-type tabular data. RVAE allows robust
identification of outliers during training, reducing their
contribution to parameter learning. Furthermore, a novel row outlier
score for mixed-type features was introduced.  RVAE outperforms or
matches competing models for OD and dirty cell repair, even though they rely
heavily on fine-tuning of hyper-parameters with a trusted labelled set.

\subsubsection*{Acknowledgements}

This work was supported by The Alan Turing Institute under the EPSRC
grant EP/N510129/1, by the EPSRC
Centre for Doctoral Training in Data Science (funded by the UK
Engineering and Physical Sciences Research Council grant
EP/L016427/1), the University of Edinburgh and in part by an Amazon Research Award. We also thank reviewers for fruitful comments and corrections. SE would like to thank Afonso Eduardo, Kai Xu 
and CUP group members for helpful discussions.

\bibliographystyle{plainnat}
\bibliography{Bib}

\end{document}


\twocolumn[

\aistatstitle{Robust Variational Autoencoders for Outlier Detection and Repair
of Mixed-Type Data}

\aistatsauthor{ Sim\~ao Eduardo$^1$\printfnsymbol{1} \And Alfredo Naz\'abal$^2$\printfnsymbol{1} \And  Christopher K. I. Williams$^{12}$ \And Charles Sutton$^{123}$}

\aistatsaddress{$^1$School of Informatics,
University of Edinburgh, UK \\ $^2$The Alan Turing Institute, UK; \hspace{0.5cm}$^3$Google Research}

]

\blfootnote{\printfnsymbol{1} Joint first authorship.}




\section{Dataset details}

\begin{table}[h]
\centering
\caption{Properties of the tabular datasets employed in the experiments.}
\begin{tabular}{|c|c|c|c|}
\hline
Dataset        & Rows & Real & Categorical \\
 & & features & features \\ \hline
Wine           & 6497   & 12              & 1                      \\ \hline
Adult          & 32561  & 5               & 10                     \\ \hline
Credit Default & 30000  & 14              & 10                     \\ \hline
Letter         & 20000  & 0               & 17                     \\ \hline
\end{tabular}
\label{t:tabular_datasets}
\end{table}

\section{Derivation of Coordinate Step for Weights}

From (6), we can write the bound $\mathcal{L}$ on $\log p(X)$ with respect to $\pi_{nd}(\x_n)$ as

\begin{eqnarray*}
\mathcal{L} & \propto & \sum_{n=1}^N \sum_{d=1}^D \pi_{nd}(\x_n)\mathbb{E}_{q_\phi(\z_n|\x_n)}[\log{p_\theta(x_{nd}|\z_n)}] \\
& + & \sum_{n=1}^N \sum_{d=1}^D \left(1-\pi_{nd}(\x_n)\right)\mathbb{E}_{q_\phi(\z_n|\x_n)}[\log{p_0(x_{nd})}] \\
& - & \pi_{nd}(\x_n)\log{\frac{\pi_{nd}(\x_n)}{\alpha}} \\
& - & (1 - \pi_{nd}(\x_n))\log{\frac{1-\pi_{nd}(\x_n)}{1-\alpha}}
\end{eqnarray*}

The derivative of this bound w.r.t. $\pi_{nd}(\x_n)$ can be easily computed:

\begin{eqnarray*}
\frac{\partial \mathcal{L}}{\partial \pi_{nd}(\x_n)} & = & \mathbb{E}_{q_\phi(\z_n|\x_n)}[\log{p_\theta(x_{nd}|\z_n)}] \\
& - & \mathbb{E}_{q_\phi(\z_n|\x_n)}[\log{p_0(x_{nd})}] \\
& - & \log{\frac{\pi_{nd}(\x_n)}{\alpha}} + \log{\frac{1-\pi_{nd}(\x_n)}{1-\alpha}}
\end{eqnarray*}

Evaluating $\frac{\partial \mathcal{L}}{\partial \pi_{nd}(\x_n)} = 0$ and solving 
for $\pi_{nd}(\x_n)$, we obtain the coordinate update for the weights:
\begin{equation*}
\hat{\pi}_{nd}(\x_n) = \frac{1}{1+\exp\left(-\left(\mathbb{E}_{q_\phi(\z_n|\x_n)}[ \log \frac{p_\theta(x_{nd}|\z_n)}{p_0(x_{nd})}] + \log{\frac{\alpha}{1-\alpha}}\right)\right)},
\end{equation*}
which is the sigmoid function applied to the expected log density ratio between the 
clean model and the outlier model plus the logit of the prior probability.

\section{Additional details for RVAE and Competing Methods}
\label{sec:supp_details_methods}

\begin{itemize}
    \item \textbf{Data Pre-Processing:} For all models and competitor methods the real
    features were standardized, i.e. subtracting by the empirical mean and dividing
    by standard deviation. One-hot encoding for categorical features was used 
    depending on the method, as defined below.
    
    \item \textbf{Validation Set}: 10\% of each dataset was separated from the rest of the data to be employed as a validation set, with known ground truth of the corrupted cells, for hyper-parameter selection on all baselines. Our RVAE model does not use this validation set in any of the 
    experiments.
    
    \item \textbf{Hyper-parameter Selection:} The criterion used for 
    hyper-parameter selection on all baselines was the AVPR in the
    outlier detection task registered in the validation set. 
    The exception is the Marginals Distribution baseline, where the number of components is
    chosen via BIC score.
    
\end{itemize}

\subsection{RVAE, VAE, DeepRPCA and Conditional Predictor methods}

\begin{itemize}

    \item \textbf{Architecture:} 
    For VAE, RVAE and DeepRPCA, we used an intermediate hidden layer in both encoder 
    and decoder, size 400. The latent space dimension was chosen to be size 20. 
    In the CondPred baseline, we found that a deep version of the base conditional predictor was
    superior than a linear one in both outlier and repair metrics. Two inner layers of dimension 
    200 and 50 for each predictor were employed, which made this model substantially slower than
    all autoencoder baselines. The non-linear activation used throughout was 
    ReLU (Rectified Linear Unit).
    
    \item \textbf{Optimization:} We used the Adam optimizer as provided in Pytorch to
    train the encoder and decoder parameters, for all VAE-based models. 
    In the case of RVAE, VAE and CondPred models we minimized 
    their respective negative losses. In CondPred, each conditional predictor had its own 
    Adam optimizer, we found this to work better.
    The initial learning rate used in experiments was $0.001$. All models ran
    for 100 epochs on all datasets, noise levels and noise processes. Since access to
    a validation set is impossible in a unsupervised learning setting, no standard early 
    stopping can be defined.
    
    In the case of DeepRPCA, we use Adam to train the encoder and decoder parameters, as 
    in the original paper. The optimization process used to obtain 
    data matrix $R$, and noise matrix
    $S$, was carried out using ADMM (Alternating Method of Multipliers).
    We use row structured $\ell_{2,1}$ version
    of DeepRPCA for outlier detection as it performed better. In order for the ADMM optimization procedure to work, in terms of categorical 
    reconstruction loss we follow the work in~\citep{udell2016generalized} (Section 6, Categorical PCA),
    using cross-entropy loss to aggregate the different one-hot dimensions. This yielded
    better experimental results than one-vs-all type aggregation.
    All models ran for 20 ADMM iterations, each 
    using 10 intermediate epochs of Adam to train the autoencoder component $R$. All the above are in 
    accordance to DeepRPCA paper \citep{zhou2017anomaly}. It should be noted that, in our experiments, running more ADMM iterations eventually led to performance degradation, even after an extensive hyper-parameter
    search and optimizer tuning.
    
    \item \textbf{$\ell_2$ Regularization (Weight Decay):} We used the weight decay option of the Adam optimizer in
    Pytorch. We performed a grid search over the values $\lambda_{\ell_2}=[0, 0.1,1,5,10,100]$, each run
    for 100 epochs, and chose the best on the validation set 
    The search was performed for each dataset in Table 1. For VAE, the best performance was obtained with we $\lambda_{\ell_2}=0.1$ in the Letter dataset, $\lambda_{\ell_2}=1$ in the Adult dataset and $\lambda_{\ell_2}=10$ in the Wine and Credit Default datasets. For the conditional predictors, the best performance was obtained for $\lambda_{\ell_2}=1$ in Adult, Credit default and Letter datasets, and $\lambda_{\ell_2}=5$ in the Wine dataset. For RVAE-CVI and RVAE-AVI no regularization was needed.
    
    \item \textbf{Categorical Encoding:} VAE, RVAE and CondPred models we used categorical embedding 
    matrices to codify the categorical features at the input level of the encoder.
    The dimensionality used in all experiments was size 50, as it provided generally
    good results. For CondPred, embeddings were not shared between individual feature predictors.
    In the case of DeepRPCA we had to use on-hot encoding, as this was the only way 
    to make the ADMM procedure to work properly, given the projection step (using
    proximity operator). This relies on subtracting the noise matrix $S$ from the
    data matrix $X$, which is non-trivial using embedding representations.
    One-hot encoding is standard in PCA-type models when dealing with categorical features.
    
    \item \textbf{DeepRPCA hyper-parameter:} The coefficient that regulates how many
    of the data-points (cells) will be represented by sparse matrix $S$ was chosen from the range $\lambda=[0.001, 0.01, 0.1, 1]$. The best outlier detection performance was obtained for 
    $0.01$ in Wine and Adult datasets and $0.1$ in Credit Default and Letter datasets.
    
    \item \textbf{RVAE (hyper-parameters):} The value for the prior probability 
    $\alpha$ was set to 0.95 throughout (it is fair to assume in general that most of the data is clean). A full evaluation on its effect on the performance of the model was conducted in the main text. In the case of the hyper-parameter $S$ of the outlier model for real features, we used $2$ throughout, with good results. This was the
    setting used for all RVAE-based models in the experiment section, and the validation set was not employed at any time while selecting parameters.
    
    \item \textbf{Encoder of the weights for RVAE-AVI:} We used a feed-forward
    neural network with the same architecture as the one specified above for the
    encoder of , which parameterizing the variational distribution of the latent space.
    An intermediate hidden layer of size 400 was used. In this case, no coordinate 
    optimization procedure was performed.
    
\end{itemize}

\subsection{OC-SVM}
We use a scikit-learn implementation, with RBF (radial basis function) kernel. We conducted an hyper-parameter search on both $\nu$ and $\gamma$, from 0 to 1 in intervals of 0.1. The best performance for all the datasets was obtained with $\nu=0.2$ and $\gamma=0.1$, on the validation set.

\subsection{Marginal Method:}
The Marginal method has no hyper-parameters to tune, apart from the maximum number 
of Gaussian Mixture Model components that can be selected by BIC score. We found
a maximum of 40 components to be sufficient.

\subsection{OCSVM + Marginals method}

We employed a combination of both the OCSVM and Marginals implementations described above. The parameters were selected based on the previous details ($\nu=0.2$,$\gamma=0.1$ and maximum number of components of the GMM set to 40).

\subsection{Isolation Forest:}
We use scikit-learn implementation. A maximum number of samples of 50\% of the size of the datasets, and a contamination parameter of 0.2 seemed to work best for all the scenarios. Again, these parameters were selected using the validation set.

\section{Outlier detection additional details}

\begin{figure}[h]
    \centering
    \includegraphics[width=1.0\linewidth]{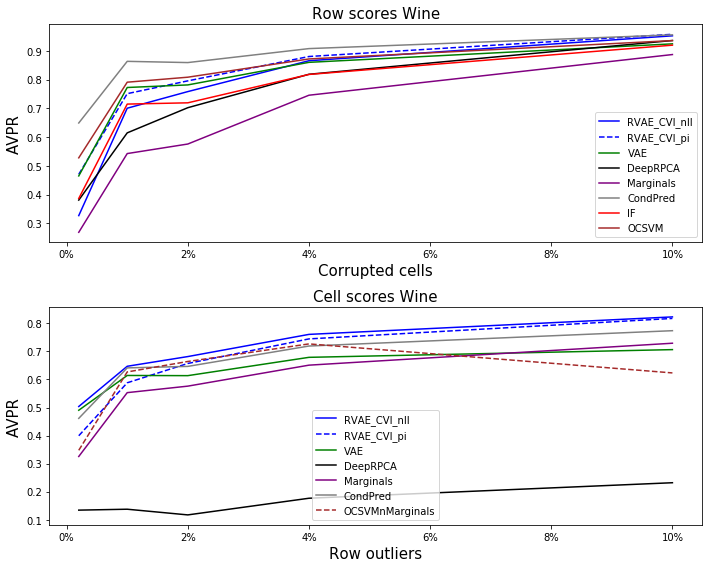}
    \vspace*{-7mm}
    \caption{Row and cell outlier detection scores on Wine dataset in 5 different cells corruption levels. Upper figure shows the AVPR at row level. Lower figure shows the AVPR at cell level.}
    \label{fig:avpr_Wine}
\end{figure}

\begin{figure}[h]
    \centering
    \includegraphics[width=1.0\linewidth]{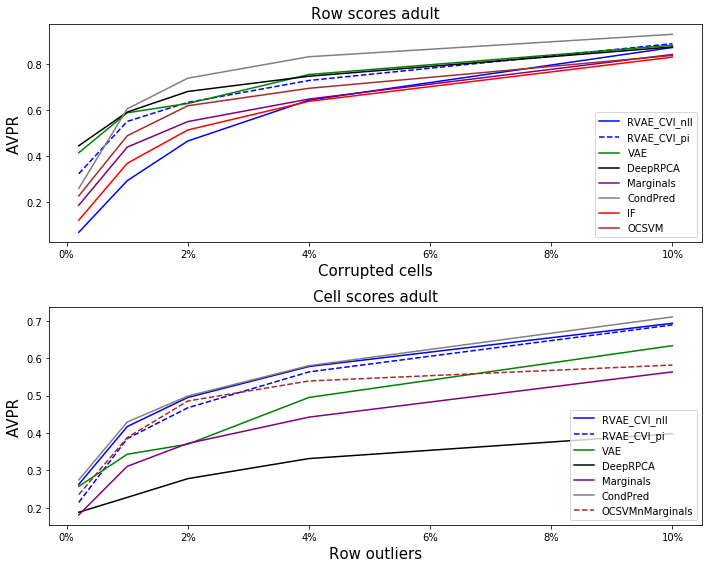}
    \vspace*{-7mm}
    \caption{Row and cell outlier detection scores on Adult dataset in 5 different cells corruption levels. Upper figure shows the AVPR at row level. Lower figure shows the AVPR at cell level.}
    \label{fig:avpr_Adult}
\end{figure}

\begin{figure}[h]
    \centering
    \includegraphics[width=1.0\linewidth]{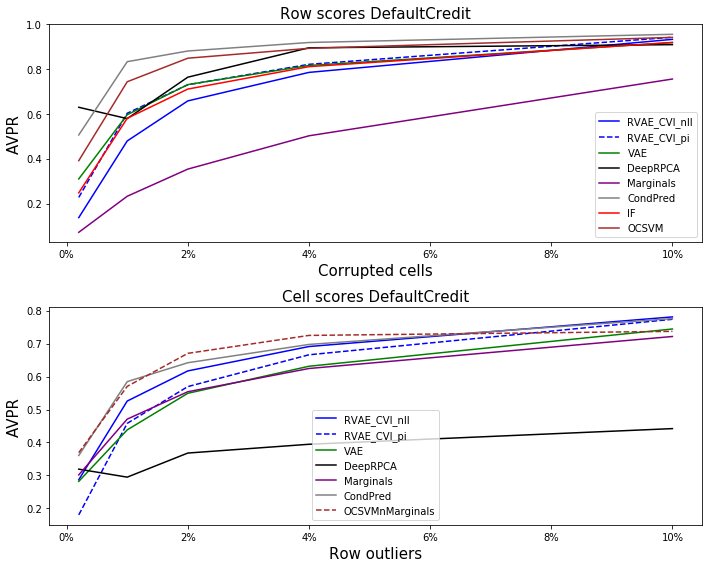}
    \vspace*{-7mm}
    \caption{Row and cell outlier detection scores on Credit default dataset in 5 different cells corruption levels. Upper figure shows the AVPR at row level. Lower figure shows the AVPR at cell level.}
    \label{fig:avpr_Default}
\end{figure}

\begin{figure}[h]
    \centering
    \includegraphics[width=1.0\linewidth]{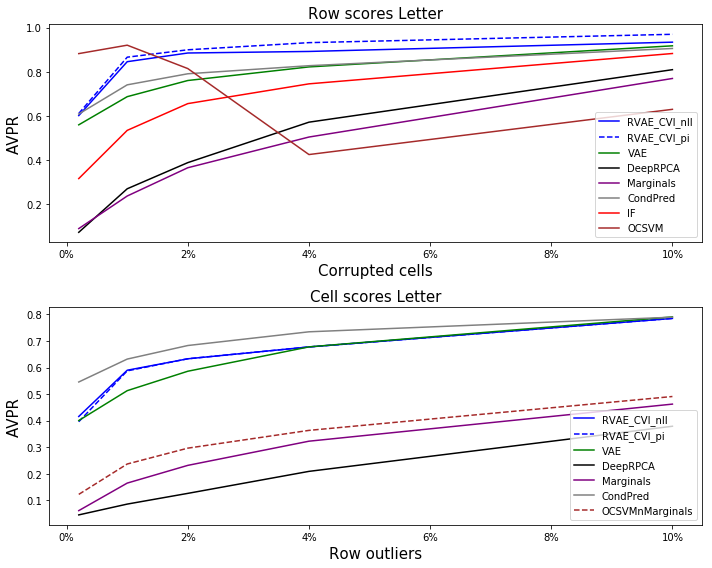}
    \vspace*{-7mm}
    \caption{Row and cell outlier detection scores on Letter dataset in 5 different cells corruption levels. Upper figure shows the AVPR at row level. Lower figure shows the AVPR at cell level.}
    \label{fig:avpr_Letter}
\end{figure}

In this section, we present the full disclosure of all the models in both row and cell outlier detection in each of the datasets of the experiments, in Figures~\ref{fig:avpr_Wine}-\ref{fig:avpr_Letter}
Notice that RVAE-CVI is stable across datasets and noise corruption levels, while other models suffer in some specific datasets for either row or cell outlier detection.

\section{Repair additional details}

\begin{figure}[h]
    \centering
    \includegraphics[width=1.0\linewidth]{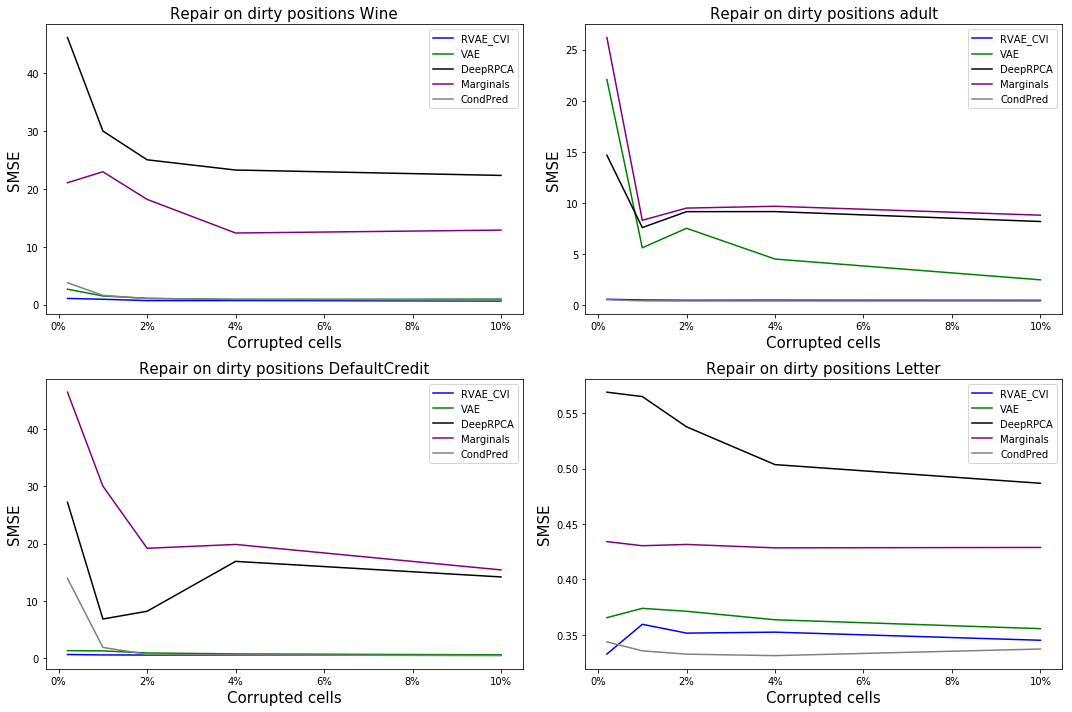}
    \vspace*{-7mm}
    \caption{Repair performance on the dirty cells of all models for each datasets}
    \label{fig:repair_datasets}
\end{figure}

In this section, we present the full disclosure of all the models in while repairing dirty cells in each of the datasets of the experiments, in Figure~\ref{fig:repair_datasets}.
RVAE-CVI performs better than the other methods for low level corruption, except for the adult dataset where RVAE-CVI and the conditional predictor are equivalent and the Letter dataset, where the conditional predictor does slightly better.

\section{RVAE-CVI vs RVAE-AVI}

We present here the AVPR evolution of RVAE-CVI and RVAE-AVI for each dataset and all noise corruption levels.
\begin{figure}[h]
    \centering
    \includegraphics[width=1.0\linewidth]{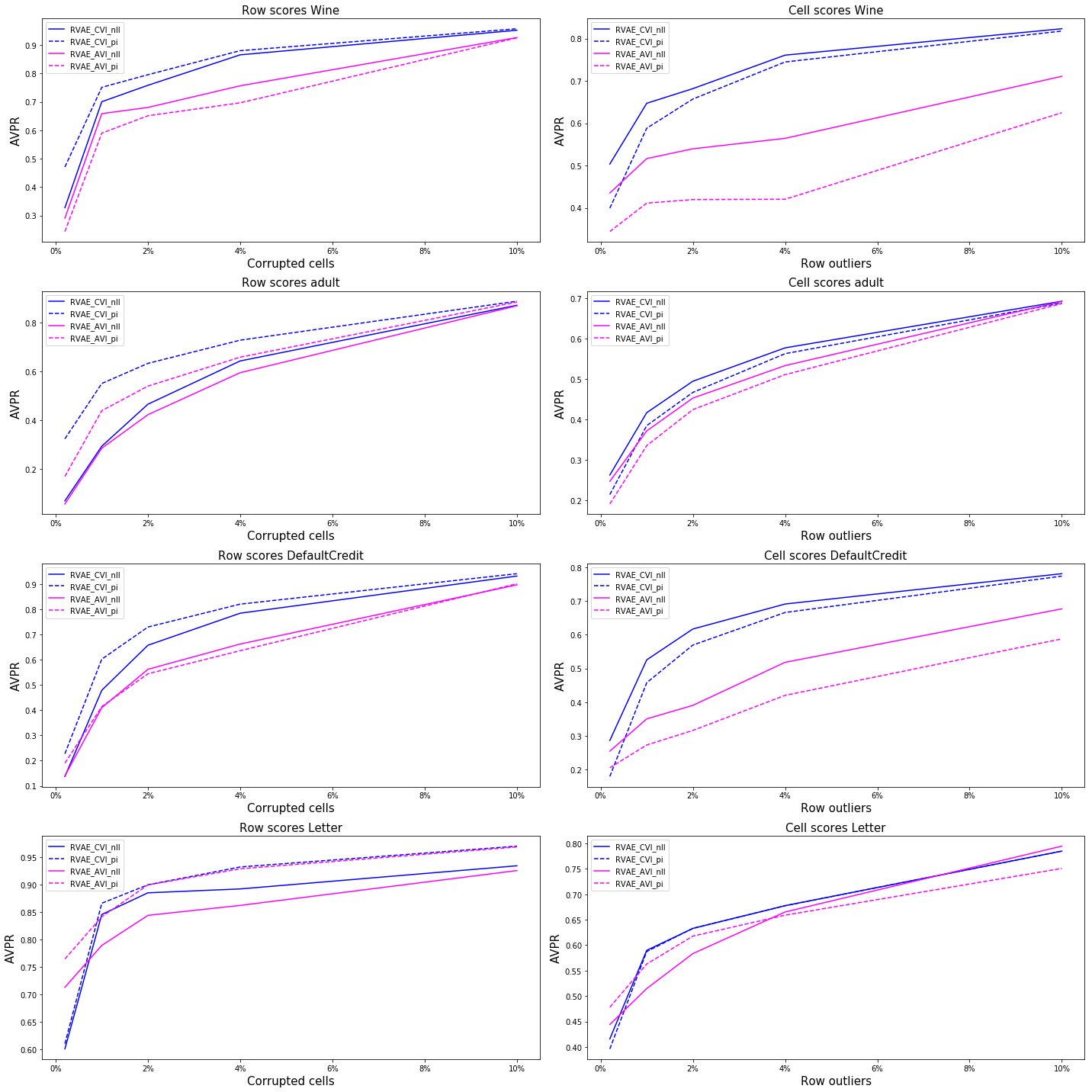}
    \vspace*{-7mm}
    \caption{Comparison between RVAE-CVI and RVAE-AVI for each dataset in row outlier detection (left figures) and cell outlier detection (right figures)}
    \label{fig:avpr_CVIvsAVI}
\end{figure}
RVAE-CVI outperforms RVAE-AVI in all datasets in both cell and row outlier detection, obtaining a similar performance only for the Letter dataset.

\begin{figure}[h]
    \centering
    \includegraphics[width=1.0\linewidth]{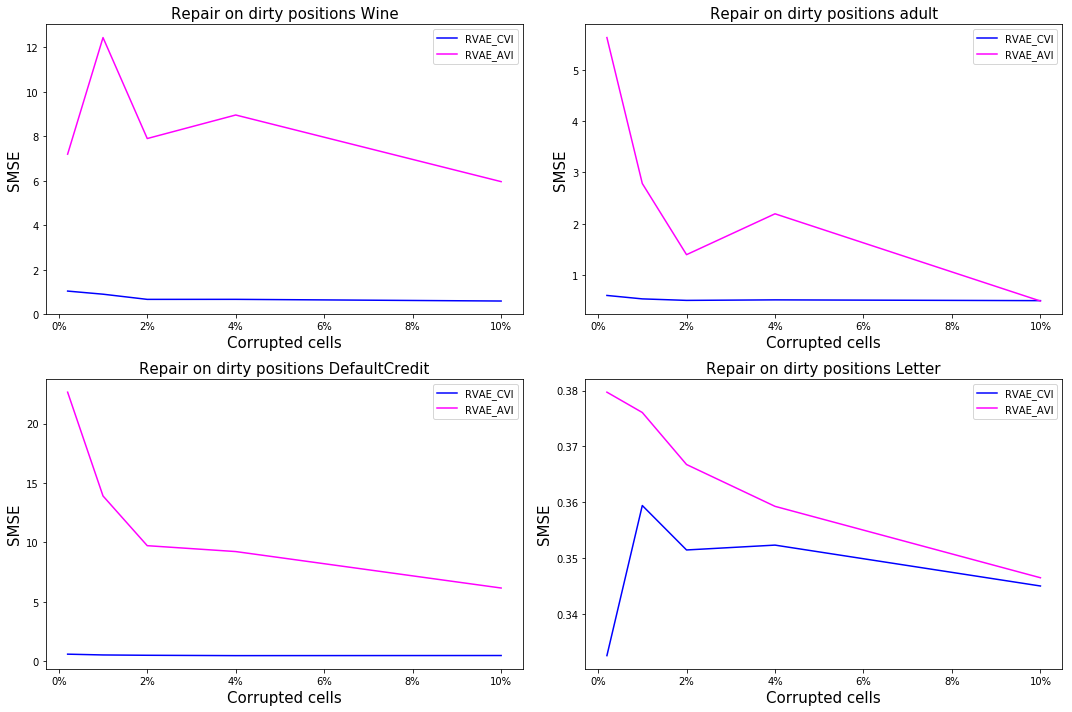}
    \vspace*{-7mm}
    \caption{Comparison between RVAE-CVI and RVAE-AVI for each dataset in repair of dirty dells. The lower SMSE the better.}
    \label{fig:repair_CVIvsAVI}
\end{figure}

Additionally, in Figure~\ref{fig:repair_CVIvsAVI} we show the difference in repair performance of the dirty cells for both models.
We can observe that RVAE-CVI performs better than RVAE-AVI for all datasets and noise corruption levels.

\section{Different noise processes additional details}

In this section we present all the results in row and cell outlier detection and repair for all six combinations of noise processes, which are:
\begin{itemize}
    \item Gaussian noise ($\mu = 0$, $\eta = 5\hat{\sigma}_d$), Tempered Categorical ($\beta = 0$)
    \item Laplace noise ($\mu=0$, $\eta = 4\hat{\sigma}_d$), Tempered Categorical ($\beta = 0.5$)
    \item Laplace noise ($\mu=0$, $\eta = 4\hat{\sigma}_d$), Tempered Categorical ($\beta = 0.8$)
    \item Laplace noise ($\mu=0$, $\eta = 8\hat{\sigma}_d$), Tempered Categorical ($\beta = 0.8$)
    \item Log normal noise ($\mu=0$, $\eta = 0.75\hat{\sigma}_d$), Tempered Categorical ($\beta = 0$)
    \item Mixture of two Gaussian noise components ($\mu_1 = -0.5, \eta_1=3\hat{\sigma}_d$, with probability 0.6 and $\mu_2 = 0.5, \eta_2=3\hat{\sigma}_d$ with probability 0.4), Tempered Categorical ($\beta = 0$)
\end{itemize}

\begin{figure}[h]
    \centering
    \includegraphics[width=1.0\linewidth]{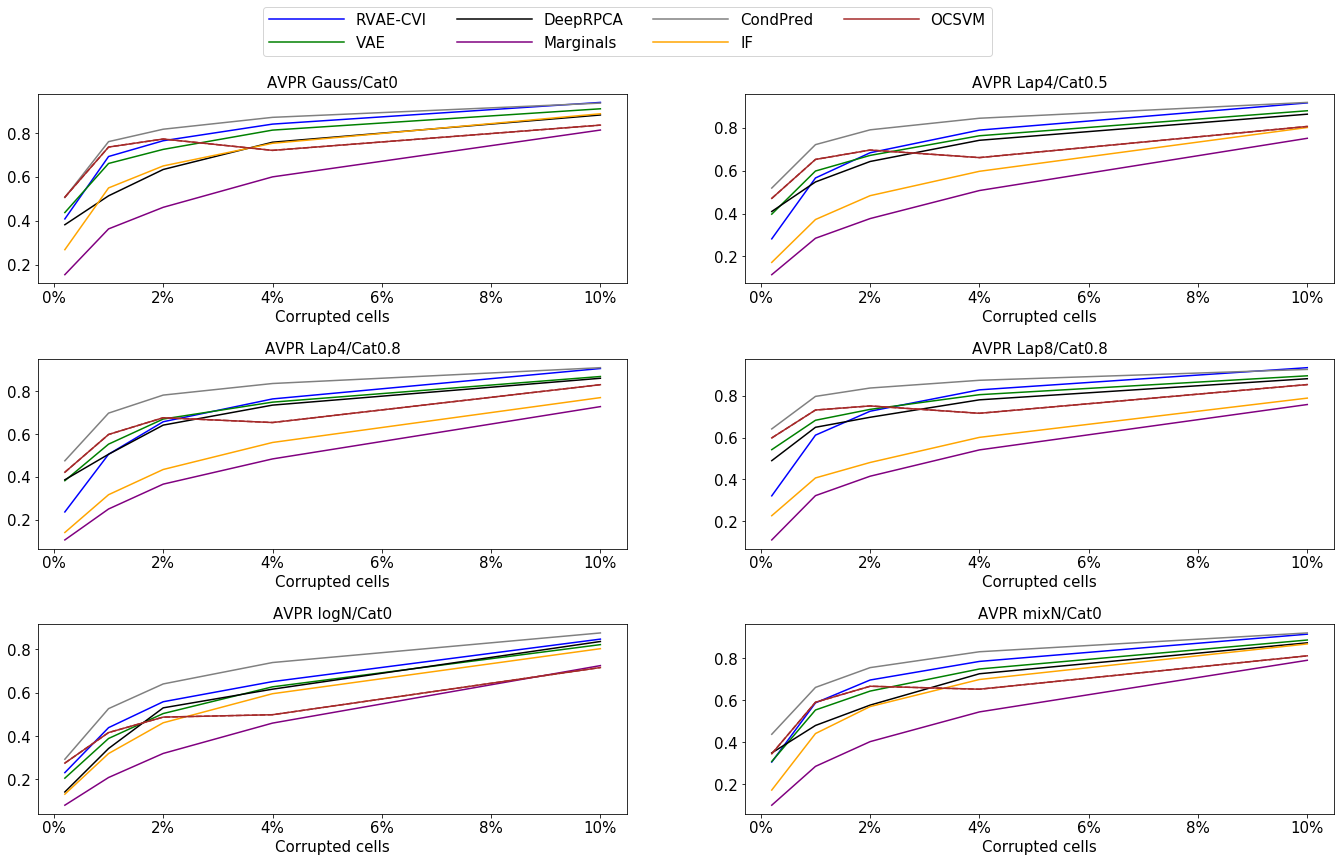}
    \vspace*{-7mm}
    \caption{Row outlier detection across all models and noise processes, averaging all datasets}
    \label{fig:noise_rowAVPR}
\end{figure}

\begin{figure}[h]
    \centering
    \includegraphics[width=1.0\linewidth]{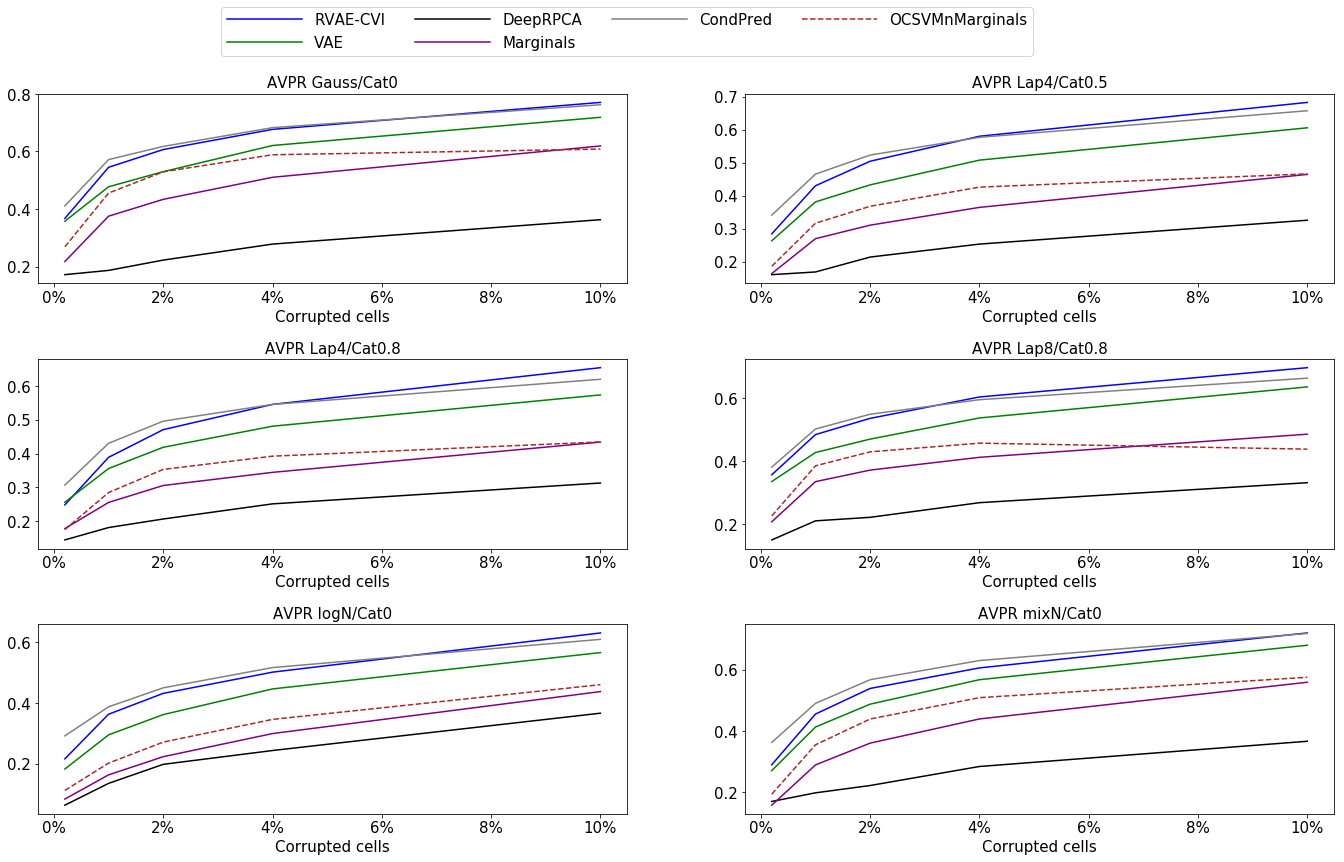}
    \vspace*{-7mm}
    \caption{Cell outlier detection across all models and noise processes, averaging all datasets}
    \label{fig:noise_cellAVPR}
\end{figure}

\begin{figure}[h]
    \centering
    \includegraphics[width=1.0\linewidth]{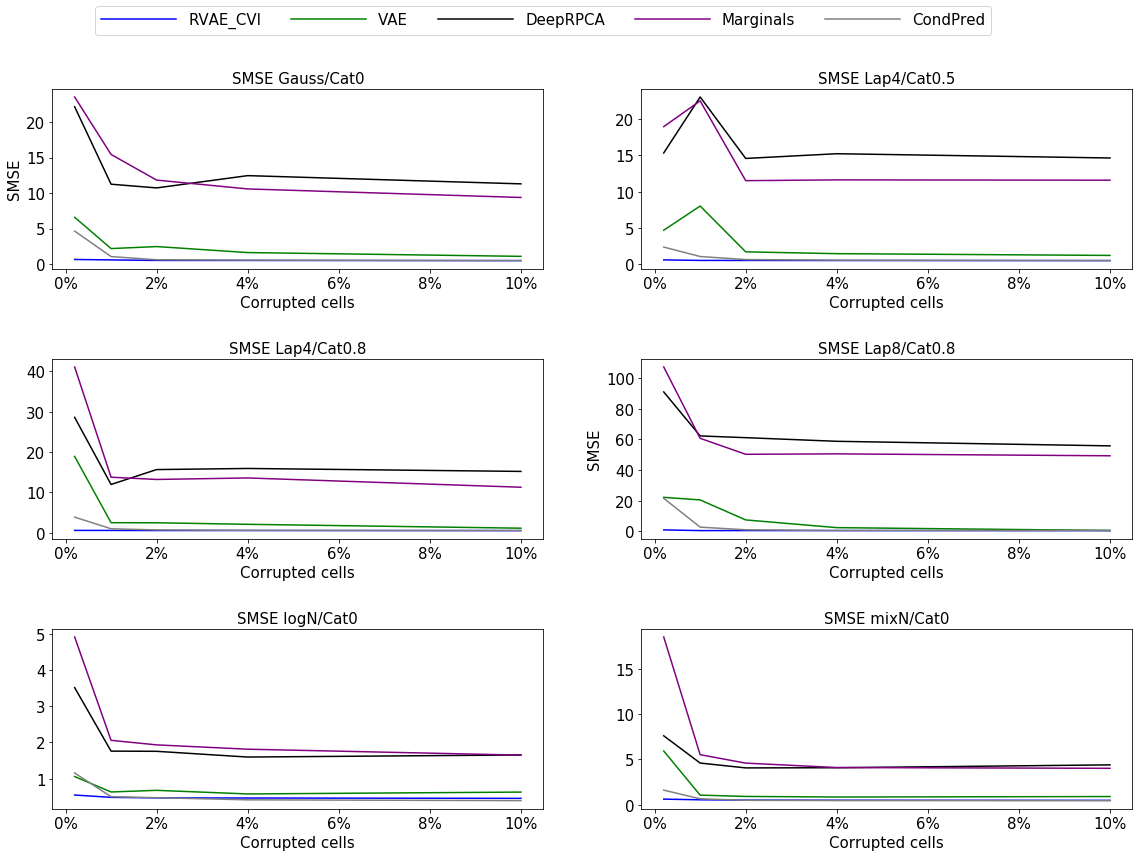}
    \vspace*{-7mm}
    \caption{Repair of dirty cells across all models and noise processes, averaging all datasets}
    \label{fig:noise_repair}
\end{figure}

Figures~\ref{fig:noise_rowAVPR}-\ref{fig:noise_repair} show a disclosure of the full results on all noise processes across the different models for both row and cell outlier detection and repair.

\section{Error Bars per Noise Level}

\begin{figure}[h]
    \centering
    \includegraphics[width=1.0\linewidth]{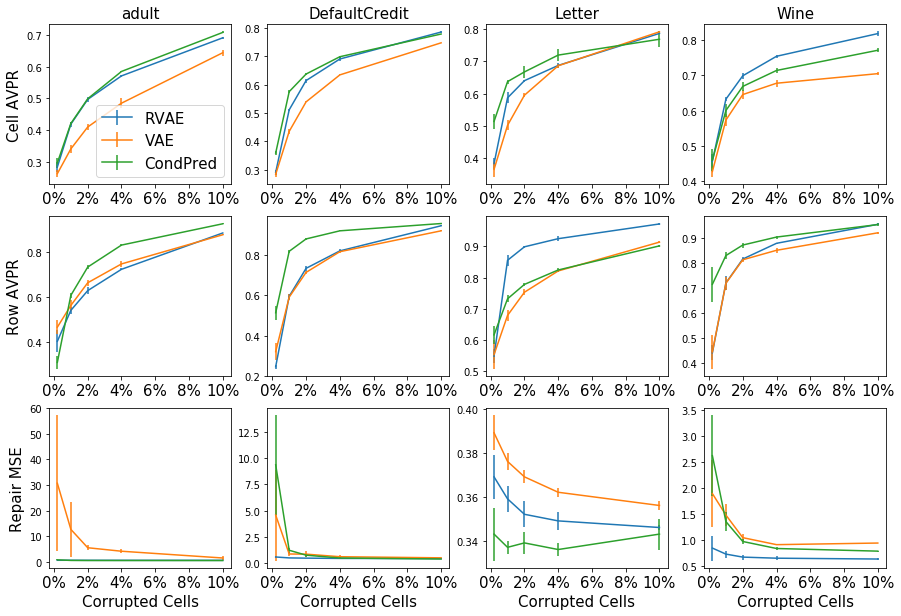}
    \vspace*{-7mm}
    \caption{Plots with error-bars for each dataset (column), using 5 different instances of
    corruption, at each corruption level (x-axis). We show cell OD (upper row), 
    row OD (middle row) and repair (lower row).}
    \label{fig:errorbars}
\end{figure}

Here, we show results for VAE, RVAE and CondPred with error bars provided for each
noise level. The error bars were obtained by generating five independent instances of corruption -- randomly corrupting different cells in the dataset each time.
The corruption process is the same as in section 4.5. of the main paper. The inference
mechanism for repair is MAP like in section 4.6. of main paper.
We report the results both for OD and repair (Figure~\ref{fig:errorbars}).
In lower noise levels, the standard deviation tends to be higher, more significantly in repair (last row, Figure~\ref{fig:errorbars}). Since fewer cells are affected at lower noise levels, this leads
to more diverse behaviours in repair and OD, and thus to larger error bar. 

We can see that the main conclusions about the "ranking" of our method against baselines still holds in either OD or repair. Further, in repair, 
in the two lowest noise levels RVAE (MAP) seems to less dependent on the corrupted cells (see Adult 
and Credit Default figures, in Figure~\ref{fig:errorbars}).

To further complete this analysis, we provide in Table~\ref{tab:pValues} the p-values computed from an independent t-test between RVAE, VAE and CondPred. These were averaged across datasets and noise levels.

\begin{table}[!]
\caption{Independent t-test between RVAE, VAE and CondPred. 
If p-values in range 0.05-0.10 assume that models have different performance.} \label{tab:pValues}
\centering
\scriptsize
\begin{tabular}{c|c|c}
 & avg. p-values &  avg. p-values\\
 & RVAE vs CondPred & RVAE vs VAE \\
\hline
Cell AVPR & 0.121 & 0.070\\
Row AVPR & 0.040 & 0.227\\
Repair SMSE & 0.025 & 0.013
\end{tabular}
\end{table}

\section{Different OD Task: RVAE vs ABDA}

In this section we compare RVAE to ABDA~\citep{vergari2019automatic}, a recent algorithm employed both in OD and missing data imputation. We followed the details in the OD section of the ABDA paper and compare RVAE with ABDA in terms 
of row AUC ROC as used therein (we use the results reported by the ABDA authors). 
\begin{table}[h]
\caption{Comparison between RVAE and ABDA in row AUC ROC for 10 different datasets.} \label{baselines}
\centering
\vspace{0.3cm}
\begin{tabular}{c|c|c}
Dataset & AUC RVAE & AUC ABDA \\
\hline
Letter &	\bf{0.8359} &	0.7036 \\
Breast &	0.9815	 &	\bf{0.9836} \\
Pen Global &	\bf{0.9316} &	0.8987 \\
Pen Local &	0.9053 &	\bf{0.9086} \\
Satellite &	\bf{0.9460} &	0.9455 \\
Thyroid &	0.8211 &	\bf{0.8488} \\
Shuttle &	\bf{0.9985} &	0.7861 \\
Aloi &	\bf{0.5515} &	0.4720 \\
Speech &	\bf{0.5584} &	0.4696 \\
KDD &	\bf{0.9993} &	0.9979 \\
\hline
Average &	\bf{0.8529} &	0.8014 \\
\hline
\end{tabular}
\label{tab:ABDA}
\end{table}
Table~\ref{tab:ABDA} shows that we perform better in average than ABDA, with 7 out 10 cases being better in OD. Notice that, the noising scenarios for these datasets (described in~\citep{goldstein2016comparative}) are based of standard row outlier detection, where one or some classes are considered normal while another class or classes are considered outliers. This scenario is completely different to the scenarios described on our paper. In our work, we assume that some cells in the data corrupt several rows in a tabular dataset, and we need to detect and correct them. These experiments showcase the robustness of RVAE to a different outlier detection process.

\section{Different Inference Method}
\label{sec:diff_infer}

\begin{figure}[h]
    \centering
    \includegraphics[width=1.0\linewidth]{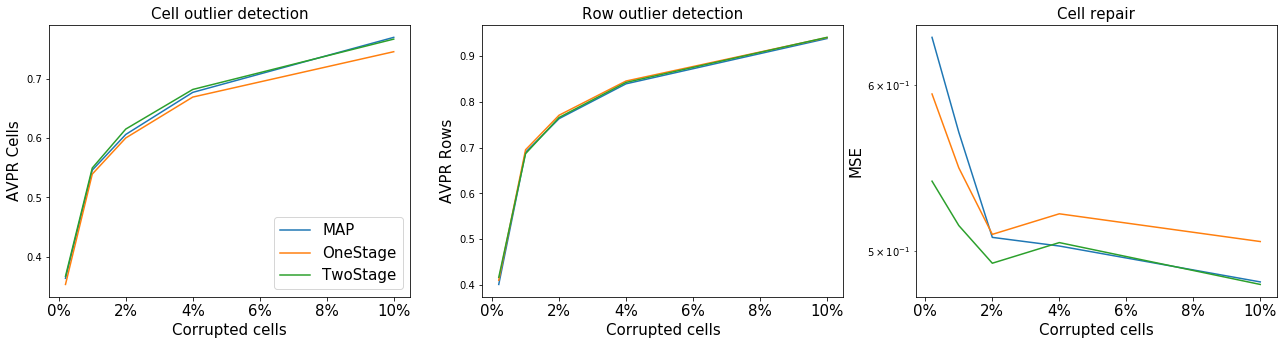}
    \vspace*{-7mm}
    \caption{Comparison between MAP, OneStage, TwoStage inference methods 
    in terms of both row / cell OD, and repair.}
    \label{fig:inference_methods}
\end{figure}

In this section, we compare the MAP inference (reconstruction, eq. (12)) for VAEs employed throughout the paper with more powerful inference methods (Figure~\ref{fig:inference_methods}). In particular, we provide results for pseudo-Gibbs sampling, (see~\citealp[section F]{rezende2014stochastic}), applying it on a trained RVAE at evaluation time. The final repair estimate was provided after running the MCMC procedure for $T=5$ iterations 
(samples), since larger values of $T$ provided marginal improvements.
We used the same scenarios of sections 4.5 and 4.6 of paper.
\begin{figure}[h]
    \centering
    \includegraphics[width=1.0\linewidth]{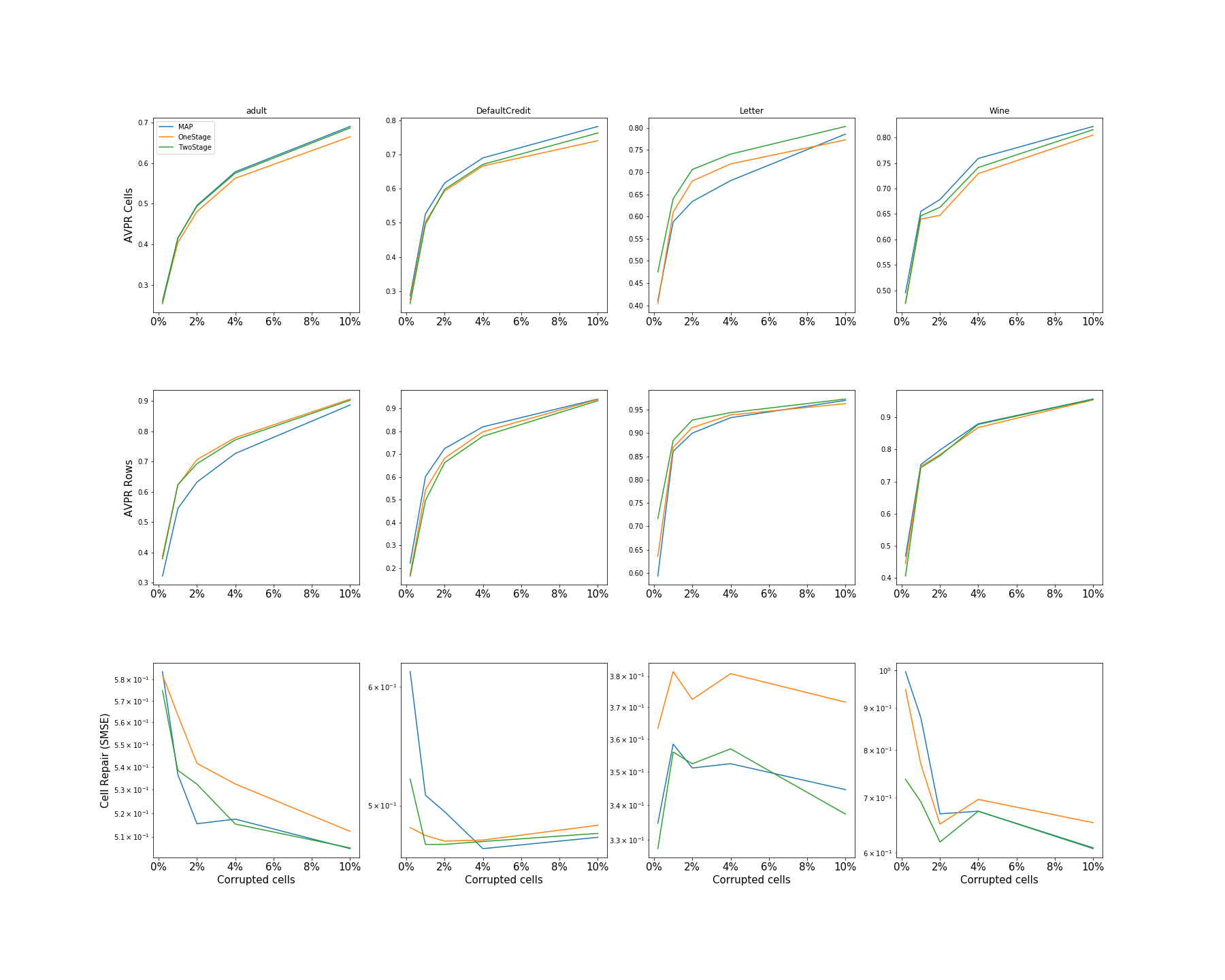}
    \vspace*{-7mm}
    \caption{Comparison between MAP, OneStage, TwoStage, CondPred inference methods 
    in terms of both row / cell OD, and repair. Results for each dataset.}
    \label{fig:inference_methods_full}
\end{figure}

A mask removing anomalous entries needs
to be either defined, or inferred. We provide two options to do this 
automatically: 
\begin{itemize}
    \item \textbf{OneStage} (Algorithm \ref{alg:one_stage}): Treat all cells in a row as anomalous and perform pseudo-Gibbs, for $T$ iterations. Final repair value is line 6 of Algorithm \ref{alg:one_stage}, 
    and OD is line 8.
    \item \textbf{TwoStage} (Algorithm \ref{alg:two_stage}): Use OneStage, obtaining a more stable estimate of $\pi_{nd}$, then sample mask $w_{nd}$ using it to perform pseudo-Gibbs (as described in~\citep{rezende2014stochastic}). 
    The assumed clean cells (i.e. $w_{nd}=1$) have their value 
    $x^o_{nd}$ fixed throughout the MCMC chain (of $T$ iterations). 
    Meanwhile cells that are dirty are initialized with mean behaviour imputation, 
    i.e. $\bar{x}_{nd}$ (Algorithm \ref{alg:two_stage}, line 4). For continuous
    features since our data is standardized ($\hat{\mu}_d=0$), so we use $0$. For categorical
    features, given our VAE models use normalized (word) embeddings, we use vectors of 
    the same dimension with zeros -- such strategy has been applied for imputation when 
    using embeddings. Final repair value is line 8 of Algorithm \ref{alg:two_stage}, 
    and OD is in line 2 (i.e. $\hat{\pi}_n$ from \textit{OneStage}).
\end{itemize}
Note that in the \textit{OneStage} method the mask $\w_n$ is not inferred, while in \textit{TwoStage} it is. In addition, we remind the reader that $\x_n^o$ is the observed row, which can be clean or dirty.

Figure~\ref{fig:inference_methods} shows that there were gains on average in OD and repair using \textit{TwoStage}, 
particularly for repair at low noise levels. 
These are still close to MAP, specifically in the case of higher noising levels.

For completion, we disclose in Figure~\ref{fig:inference_methods_full} the comparison across the inference methods per dataset.
In general, we can see that TwoStage has better repair performance (last row of Figure~\ref{fig:inference_methods_full}), particularly in low level noise. 

Lastly, other methods like \citep{Mattei2019MIWAEDG} could also have been used to improve repair. 
However, more powerful inference schemes can sometimes lead to overfitting to noise. 
On the other hand, inference schemes like MCMC (vs MAP) can provide more stable 
solutions (lower error bars), particularly in lower noise levels or 
in smaller datasets (number of rows).

\begin{algorithm}[ht]
\caption{OneStage: pseudo-Gibbs sampling}
\label{alg:one_stage}
\begin{algorithmic}[1]
\Procedure{OneStage}{$T$, fixed \{$\phi$, $\theta$\}}
    \State $\x_n^{(1)} = \x_n^o$ 
    \For{$1,...,T$}
        \State $\z_n^{(t+1)} \sim q_\phi(\z|\x_n^{(t)})$
        \State $\tilde{\x}_n^{(t+1)} \sim p_\theta(\x|\z_n^{(t+1)})$
    \EndFor
    \State $\hat{\x}^i_n = \tilde{\x}_n^{(T+1)}$ \Comment{for repair}
    \State $\hat{\z}^i_n = \z_n^{(T+1)}$
    \State $\hat{\pi}_{nd} = g\left(r(\x_{nd}^o, \hat{\z}^i_n) + \log \frac{\alpha}{1-\alpha}\right)$ \Comment{eq.(9), OD}
    \State \Return $(\hat{\x}^i_n, \hat{\z}^i_n, \hat{\pi}_n)$
\EndProcedure
\end{algorithmic}
\end{algorithm}

\begin{algorithm}[ht]
\caption{TwoStage: pseudo-Gibbs sampling}
\label{alg:two_stage}
\begin{algorithmic}[1]
\Procedure{TwoStage}{$T$, fixed \{$\phi$, $\theta$\}}
    \State $(\hat{\x}^i_n, \hat{\z}^i_n, \hat{\pi}_n)$ $\leftarrow$ OneStage($T$,\{$\phi$, $\theta$\}) 
    \State $\hat{w}_{nd} \sim q_{\hat{\pi}_{nd}}(w_{nd})$ 
    \State $x_{nd}^{(1)} = \hat{w}_{nd} \times x_{nd}^o  + (1-\hat{w}_{nd}) \times \bar{x}_{nd}$ 
    \For{$1,...,T$}
        \State $\z_n^{(t+1)} \sim q_\phi(\z|\x_n^{(t)})$
        \State $\tilde{\x}_n^{(t+1)} \sim p_\theta(\x_n|\z_n^{(t+1)})$
    \EndFor
    \State $\hat{x}^i_{nd} = \hat{w}_{nd} \times x_{nd}^o  + (1-\hat{w}_{nd}) \times \tilde{x}_{nd}^{(T+1)}$ 
    \State $\hat{\z}^i_n = \z_n^{(T+1)}$
    \State \Return $(\hat{\x}^i_n, \hat{\z}^i_n, \hat{\pi}_n)$
\EndProcedure
\end{algorithmic}
\end{algorithm}

\vfill

\bibliographystyle{plainnat}
\bibliography{Bib}